\documentclass[journal]{IEEEtran}

\usepackage{amssymb}

\usepackage[english]{babel}
\usepackage[utf8]{inputenc}
\usepackage{tabularx,caption}
\usepackage{graphicx}
\usepackage{url}
\usepackage{algorithm}
\usepackage{algpseudocode}
\usepackage{listings}
\usepackage{amsmath}
\usepackage{amssymb}
\usepackage{array}
\usepackage{float}
\usepackage[table]{xcolor}
\usepackage{multirow}
\usepackage{hhline}
\usepackage{fancyhdr}
\usepackage{units}
\usepackage{pgfplots}
\usepackage{rotating} 
\pgfplotsset{compat=1.9}

\usetikzlibrary{patterns}

\usepackage[draft]{hyperref}
\hypersetup{
	bookmarks=true,        
    unicode=true,          
    pdftoolbar=true,        
    pdfmenubar=true,        
    pdffitwindow=false,     
    pdfstartview={FitH},    
    pdftitle={Monte Carlo Tree Search algorithms applied to the card game Scopone},    
    pdfauthor={Stefano Di Palma},     
    pdfsubject={Master's Thesis},   
    pdfkeywords={MCTS} {Scopone} {Scopone scientifico} {ISMCTS} {AI} {Monte Carlo Tree Search} {Information Set Monte Carlo Tree Search} {Artificial Intelligence}, 
    pdfnewwindow=true,      
    colorlinks=false,       
    hidelinks,
    linkcolor=red,          
    citecolor=green,        
    filecolor=magenta,      
    urlcolor=cyan           
}

\newcounter{rule}[section]

\makeatletter
\let\OldStatex\Statex
\renewcommand{\Statex}[1][3]{%
  \setlength\@tempdima{\algorithmicindent}%
  \OldStatex\hskip\dimexpr#1\@tempdima\relax}

\def\namedlabel#1#2{\begingroup
	\stepcounter{rule}
    #2\arabic{rule}%
    \def\@currentlabel{#2\arabic{rule}}%
    \phantomsection\label{#1}\endgroup
}
\makeatother

\begin{document}

\title{Traditional Wisdom and Monte Carlo Tree Search Face-to-Face in the Card Game Scopone}


\author{Stefano Di Palma and Pier Luca Lanzi\\
Politecnico di Milano --- Dipartimento di Elettronica, Informazione e Bioingegneria\\
\thanks{Corresponding author: Pier Luca Lanzi (email: pierluca.lanzi@polimi.it).}
}

%
%
%

\maketitle

\newdimen\plotheight
\plotheight=6cm
\newdimen\plotwidth
\plotwidth=\columnwidth

\newcommand{\Scopone}{\emph{Scopone}}
\newcommand{\EGS}{$\epsilon$GS}

\begin{abstract}
We present the design of a competitive artificial intelligence for \emph{Scopone}, a popular Italian card game.
We compare rule-based players using the most established strategies (one for beginners and two for advanced players)
	against players using Monte Carlo Tree Search (MCTS) and Information Set Monte Carlo Tree Search (ISMCTS) with 
	different reward functions and simulation strategies.
MCTS requires complete information about the game state and thus implements a cheating player 
   while ISMCTS can deal with incomplete information and thus implements a fair player.
Our results show that, as expected, the cheating MCTS outperforms all the other strategies;
	 ISMCTS is  stronger than all the rule-based players implementing well-known and most advanced strategies 
	 and it also turns out to be a challenging opponent for human players.
\end{abstract}



\section{Introduction}
\textit{Scopone} is a popular Italian card game whose origins date back to (at least) 1700s.
Originally, the game was played by poorly educated people and
	as any other card game in Italy at the time was hindered by  authority. 
Later,
	being considered intellectually challenging, the game spread among highly educated people and high rank politicians\footnote{E.g., \url{http://tinyurl.com/ThePresidentAndScopone}}
achieving, nowadays, a vast popularity and a reputation similar to Bridge (for which  
	it is often referred to as \emph{Scopone Scientifico}, that is, Scientific Scopone). 
The first known book about Scopone was published in 1750 by Chitarella \cite{chitarrella2002regole} and contained both
	the rules of the game and a compendium of strategy rules for advanced players.
Unfortunately,	there are no copies available of the \textit{original} book and the eldest reprint is from 1937. 
The first original book about Scopone, that is still available, 
	was written by Capecelatro in 1855, ``Del giuoco dello Scopone''.\footnote{\url{http://tinyurl.com/Scopone}}
At that time, Capecelatro wrote that the game was known by 3-4 generations, therefore it might have been born in the eighteenth century; 
	however several historians argue that \Scopone\ was known centuries before Chitarella \cite{saracino2011scopone}.
Although dated, the rules by Chitarella \cite{chitarrella2002regole} are still considered the main and most important strategy guide for Scopone;
	only recently Saracino \cite{saracino2011scopone}, a professional bridge player, proposed additional rules to extend the original strategy 
\cite{chitarrella2002regole}.
The second most important strategy book was written by Cicuti and Guardamagna~\cite{cicuti1978segreti} who enriched \cite{saracino2011scopone} 
	by introducing advanced rules regarding the play of sevens.

In this paper, we present the design of a competitive artificial intelligence for Scopone and 
	we compare the performance of three rule-based players that implement well-established playing strategies against 
	players based on Monte Carlo Tree Search (MCTS) \cite{DBLP:conf/cg/Coulom06}
	and Information Set Monte Carlo Tree Search (ISMCTS) \cite{DBLP:journals/tciaig/CowlingPW12}.
The first rule-based player implements the basic greedy strategy taught to beginner players; 
the second one implements Chitarella's rules \cite{chitarrella2002regole} with the additional rules introduced by Saracino \cite{saracino2011scopone},
	and represents the most fundamental and advanced strategy for the game;
the third rule-based player extends the previous approach with the additional rules introduced in \cite{cicuti1978segreti}.
MCTS requires the full knowledge of the game state (that is, of the cards of all the players) 
	and thus, by implementing a cheating player, it provides an upper bound to the performance achievable with this class of methods.
ISMCTS can deal with incomplete information and thus implements a fair player.
For both approaches, 
	we evaluated different reward functions and simulation strategies.
We performed a set of experiments to select the best rule-based player and the best configuration for MCTS and ISMCTS.
Then, we performed a tournament among the three selected players and also an experiment involving humans.
Our results show that 
	the cheating MCTS player outperforms all the other strategies while the fair ISMCTS player
	outperforms all the rule-based players that implement 
	the best known and most studied advanced strategy for Scopone.
The experiment involving human players suggests that ISMCTS might be more challenging than traditional strategies.


\section{Monte Carlo Tree Search}
\label{sec:mcts2}
Monte Carlo Tree Search (MCTS) identifies a family of decision tree search algorithms that has been successfully applied to a wide variety of games \cite{DBLP:journals/tciaig/BrownePWLCRTPSC12} ranging from classic board games \cite{fuego,hex}, modern board games or Eurogames\footnote{\url{https://en.wikipedia.org/wiki/Eurogame}} \cite{scotlandYard,DBLP:conf/acg/SzitaCS09}, video games \cite{MCTSTotalWar2014,MCTSFableLegends}, 
General Video Game Playing \cite{DBLP:conf/cig/FrydenbergART15,DBLP:conf/ijcai/KhalifaITN16,DBLP:conf/cig/Nelson16}, and card games 
\cite{ponsen2010integrating,DBLP:journals/corr/Walton-RiversWB17,magic,DBLP:journals/tciaig/CowlingPW12}.

MCTS algorithms iteratively build partial and asymmetric decision trees by performing several random simulations. 
The basic MCTS algorithm comprises four steps: (i) selection, (ii) expansion, (iii) simulation, and (iv) backpropagation.

\medskip\noindent\textbf{Selection:} 
	a tree policy is used to descend the current search tree by selecting the most urgent node to explore until either a leaf 
	(representing a terminal state) or a not fully expanded node is reached; the tree policy typically uses 
	the Upper Confidence Bound for Trees (UCB1 for Trees or UCT) \cite{Auer2002235,DBLP:conf/ecml/KocsisS06}
	to descend the child node that maximizes the UCT value computed as,
\begin{equation}	
UCT(v') = \frac{Q(v')}{N(v')} + c\sqrt{2\frac{\ln N(v)}{N(v')}}
\label{eq:uct}
\end{equation}
where $v$ is the parent node, $v'$ is a child node, $N(v)$ is the number of times $v$ (the parent of $v'$) has been visited, 
	$N(v')$ is the number of times the child node $v'$ is visited, 
	$Q(v')$ is the total reward of all playouts that passed through $v'$, and $c > 0$ is a constant that controls the amount of exploration.
	
\medskip\noindent\textbf{Expansion:} 
	if the node does not represent a terminal state, then one or more child nodes are added to expand the tree.
	A child node represents a state reached by applying an available action to the current state.

\medskip\noindent\textbf{Simulation:} 
	a simulation is run from the current position in the tree to evaluate the game outcome by following a default policy (typically random);
	this results in a reward computed according to a \emph{reward function}.

\medskip\noindent\textbf{Backpropagation:} 
	the reward received is finally backpropagated to the nodes visited during selection and expansion to update their statistics.

\medskip
Cowling et al. \cite{DBLP:journals/tciaig/CowlingPW12,DBLP:conf/cig/WhitehousePC11} introduced \textit{Information Set Monte Carlo Tree Search} (ISMCTS) to extend
	MCTS to decisions with imperfect information about the game state and performed actions.
ISMCTS builds asymmetric search trees of information sets containing game states that are indistinguishable from the player's point of view.
Each node represents an information set from the root player's point of view and arcs correspond to moves played by the corresponding player.
The outgoing arcs from an opponent's node represent the union of all moves available in every state within that information set, because the player cannot know the moves that are really available to the opponent.
The selection step is changed because the probability distribution of moves is not uniform 
	and a move may be available only in fewer states of the same information set.
Thus, at the beginning of each iteration, a determinization is sampled from the root information set and 
	the following steps of the iteration are restricted to regions that are consistent with that determinization.
The probability of a move being available for selection on a given iteration is precisely the probability of sampling a determinization in which that move is available.
The set of moves available at an opponent's node can differ between visits to that node, hence UCT (Eq. \ref{eq:uct}) is modified as \cite{DBLP:conf/cig/WhitehousePC11},
\begin{equation}	
ISUCT(v') = \frac{Q(v')}{N(n')} + c\sqrt{\frac{\ln N'(v')}{N(v')}}
\label{eq:uct-ismcts}
\end{equation}	

\noindent
where $N'(v')$ is the number of times the current node $v$ (the parent of $v'$) has been visited and node $v'$ was available for selection.
Cowling et al. \cite{DBLP:journals/tciaig/CowlingPW12,DBLP:conf/cig/WhitehousePC11} introduced three ISMCTS algorithms:
Single-Observer ISMCTS (SO-ISMCTS) deals with problems involving partial information about the current state;
SO-ISMCTS With Partially Observable Moves (SO-ISMCTS+POM) extends SO-ISMCTS to the case of partially observable moves;
Multiple-Observer ISMCTS (MO-ISMCTS) improves opponents modeling by maintaining separate trees for each player searching them simultaneously.


\section{Related Work}
\label{sec:related}
Card games are challenging testbeds for artificial intelligence algorithms involving computationally complex tasks \cite{baffier_et_al:LIPIcs:2016:5864}, imperfect information, and multi-player interactions.
\emph{Poker} has been extensively studied in this context with an annual competition run since 2006 at AAAI.\footnote{\url{http://www.computerpokercompetition.org}}
In addition to imperfect information and non-cooperative multi-playing, the central element in Poker is the psychological factor of the \emph{bluff}. 
During the 2017 competition, Brown and Sandholm's Libratus \cite{Sandholm2017,CMUNews} (the successor of Claudico \cite{claudico}) has achieved an historical result by beating four top-class human poker players. 
Rubin and Watson \cite{DBLP:journals/ai/RubinW11} provide a recent overview of the field. More information is available in proceedings of the Computer \textit{Poker and Imperfect Information Games} workshop that is held during the annual competition \cite{DBLP:conf/aaai/2016poker,DBLP:conf/aaai/2015poker}. 

\textit{Bridge} also provided many ideas for research in this field \cite{wiki:BrigeGioco} with a Computer Bridge championship running since 1997.\footnote{\url{https://bridgerobotchampionship.wordpress.com/}}
In the early '80s Throop et al. \cite{bridge} developed the first version of \emph{Bridge Baron} which used a planning technique called \emph{hierarchical task network} (HTN) \cite{BridgeBaron}, based on the decomposition of tasks~\cite{BridgeBaron}.
In 1999, Ginsberg et al. \cite{IArussel} developed a program called GIB (Ginsberg's Intelligent Bridgeplayer), which won the world championship of computer bridge in 2000.
Over the years, stronger players have been developed including Jack\footnote{\url{http://www.jackbridge.com/}} and Wbridge5\footnote{\url{http://www.wbridge5.com/}}; a short survey of some of these programs is available in \cite{BridgeSoftwareHistory}.

\textit{Mahjong} has also been used as a testbed. In 2009, Wan Jing Loh \cite{mahjong} developed an artificial intelligence to play \emph{Mahjong}.  More recently, Mizukami and Tsuruoka \cite{DBLP:conf/cig/MizukamiT15} developed an approach to build a \textit{Mahjong} program that models opponent players and performs Monte Carlo simulations with the models. Their program achieved a rating that was significantly higher than that of the average human players.

Monte Carlo Tree Search (MCTS) attracted the interest of researchers in this area due to the results obtained with \emph{Go}~\cite{fuego}.
However, MCTS cannot deal with \emph{imperfect information}, therefore it requires the integration with other techniques to play typical card games.
For example, Ponsen et al. \cite{ponsen2010integrating} integrated MCTS with a Bayesian classifier, which is used to model the behavior of the opponents, to develop a \emph{Texas Hold'em Poker} player.
The Bayesian classifier is able to predict both the cards and the actions of the other players. 
Ponsen's program was stronger than rule-based artificial intelligence, but weaker than the program Poki \cite{IApoker}.
In 2011, Nijssen and Winands~\cite{scotlandYard} used MCTS in the artificial intelligence of the board game \emph{Scotland Yard}.
In this game the players have to reach with their pawns a player who is hiding on a graph-based map.
The escaping player shows his position at fixed intervals, the only information that the other players can access is the type of location (called \emph{station}) where they can find the hiding player.
In this case, MCTS was integrated with \emph{Location Categorization}, a technique
which provides a good prediction on the position of the hiding player.
Nijssen and Winands showed that their program was stronger than the artificial intelligence of the game Scotland Yard for Nintendo DS, 
	considered to be one of the strongest player.
Whitehouse et al. \cite{DBLP:conf/cig/WhitehousePC11} analyzed the strengths and weaknesses of determinization coupled with MCTS on Dou Di Zhu, 
	a popular Chinese card game of imperfect information, and introduced Information Set UCT.

In 2012, 
Cowling et al. \cite{magic} applied MCTS with determinization approaches on a simplified variant of the game \emph{Magic: The Gathering}.
They later extended the approach \cite{DBLP:journals/tciaig/CowlingPW12} 
	and introduced \emph{Information Set Monte Carlo Tree Search} (ISMCTS)
	an extension of MCTS to games with partial information on the user state and/or on the users' actions.
Cowling el al. \cite{spades,DBLP:journals/tciaig/CowlingDPWR15} applied ISMCTS to the design of an artificial intelligence for \emph{Spades}, a four players card game.
The program demonstrated excellent performance in terms of computing time.
Sephton et al. \cite{DBLP:conf/cig/SephtonCPS14} investigated move pruning in Lords of War and later
 compared different selection mechanism for MCTS to produce more entertaining opponents in the same game.
Robilliard et al. \cite{DBLP:conf/ecai/RobilliardFT14} applied MCTS using determinization to \textit{7 Wonders} a well-known card game 
involving partially observable information, multiplayer interactions, and stochastic outcomes. They showed 
convincing results, both against a human designed rule-based artificial player and against experienced human players.
Cowling et al. \cite{DBLP:conf/cig/CowlingWP15} extended MCTS with mechanisms for performing inference and bluffing and tested the algorithm
on \textit{The Resistance}, a card game based around hidden roles.
%
%
%
%

Recently, Walton-Rivers et al. \cite{DBLP:journals/corr/Walton-RiversWB17} 
	studied agent modeling in the card game \textit{Hanabi}\footnote{\url{https://en.wikipedia.org/wiki/Hanabi_(card_game)}}
	and compared rule-based players with an ISMCTS player, which showed poor performance. Accordingly they 
	developed a \emph{predictor} player using a model of the players with which it is paired and show 
	significant improvement in game-playing performance in comparison to ISMCTS.
Canezave \cite{Cazenave2006} developed {\sc Illusion}, a Phantom Go player for a 9x9 board, that uses Monte-Carlo simulations to reach 
	the level of experienced Go players who only played few Phantom Go games.
Later, Borsboom et al. \cite{Borsboom2007} presented a rather comprehensive comparison of several Monte-Carlo based players for Phantom Go
	using both simulation and tree-search. 
Ciancarini and Favini \cite{CIANCARINI2010670} compared the best available minimax-based Kriegspiel (or \emph{invisible chess}) program to three MCTS approaches. Their results show better performance with less domain-specific knowledge. 
Recently, Wang et al. \cite{7317917} presented an extension of Monte Carlo Tree Search in which nodes are belief states. The Belief-State MCTS (BS-MCTS) has an additional initial sampling step and uses belief states also during selection and when the opponent strategy is considered. 

\section{Scopone}
\label{sec:scopone}
In this section, we provide a brief overview of the rules of Scopone 
	and refer the reader to \cite{chitarrella2002regole,saracino2011scopone} for 
	a detailed description and playing strategies.

\subsection{Cards}
\label{sec:ScoponeCards}
\begin{figure}[tb]
\centering
\includegraphics[width=\columnwidth]{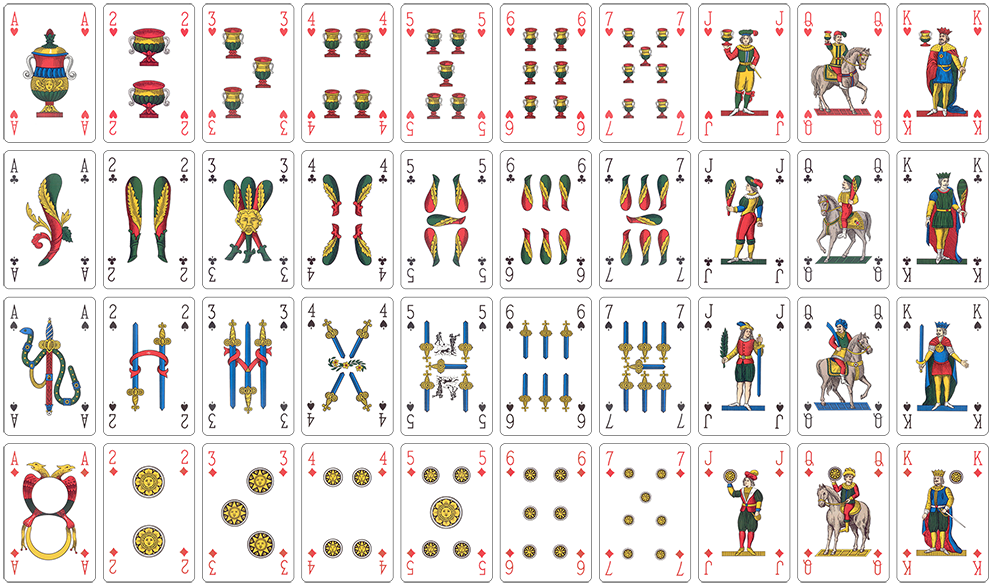}
\caption{Official deck of Federazione Italiana Gioco Scopone.
\label{fig:figsCards}}
\end{figure}

Scopone is played with the traditional Italian deck composed by 40 cards, divided in four suits 
	(\emph{coins} or \textit{Denari}, \emph{swords} or \textit{Spade}, \emph{cups} or Coppe, and \emph{batons} or \textit{Bastoni}).
Each suit contains, in increasing order, an ace (or one), numbers two through seven, and three face cards:
the \emph{knave} (Fante or eight); 
the \emph{horse} (Cavallo or nine) or \emph{mistress} (Donna or nine); 
and the \emph{king} (Re or ten).
The deck is not internationally well-known as the 52 cards French deck, probably because each region of Italy has its own cards style.
Accordingly, an official deck (Figure~\ref{fig:figsCards}) that merges the Italian and French suits was later introduced (Figure~\ref{fig:figsCards})
  in which \textit{coins} correspond to diamonds ($\diamondsuit$), 
  \textit{swords} to spades ($\spadesuit$), \textit{cups} to hearts ($\heartsuit$), and \textit{batons} to clubs ($\clubsuit$).

\subsection{Game Rules}
\label{sec:ScoponeRules}
Scopone is played by four players divided into two teams.
The players are positioned in the typical North-East-South-West positions and teammates are in front of each other.
At the right of the dealer there is the \emph{eldest hand} (in Italian ``primo di mano''), followed by the \emph{dealer's teammate} and then by the \emph{third hand}.
The team of the dealer is called \emph{deck team}, whereas the other is called \emph{hand team}.
A game consists of one or more matches and each match consists of nine rounds; a game ends when a team reaches a target score (either 11, 16, 21 or 31).
If both the teams reach the target  score, then the game continues until one team scores more than the other one.

\subsection{Cards Distribution}
\label{subsec:dealer}
\begin{figure}
\centering
\includegraphics[width=\columnwidth]{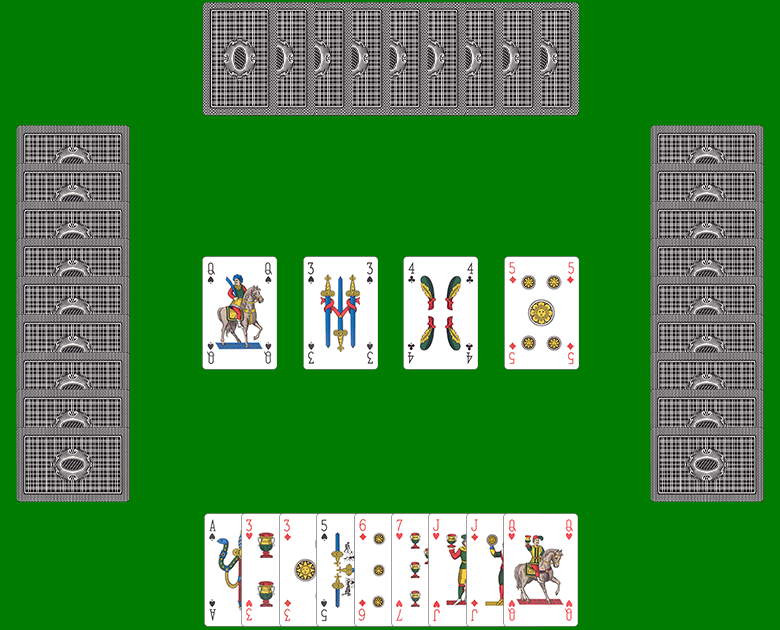}
\caption{Initial configuration of a match. Each player holds nine cards and there are four cards on the table.
\label{fig:initialGame}}
\end{figure}

Initially, the dealer is randomly chosen through some procedure, because being the last player to play has some advantages \cite{chitarrella2002regole}.
At the beginning of each match, the dealer shuffles the deck and offers it to the third hand for cutting.
Then, it deals the cards counterclockwise three by three, starting with the eldest hand (the person to the right of the dealer), for a total of nine cards for each player.
The dealer must not reveal the cards, if this happens it must repeat the process.
During the first distributions, the dealer also leaves twice on the table a pair of face-up cards, for a total of four cards.
If at the end of the distribution on the table there are three kings, the dealer must repeat the process.
At the end of the distribution the table looks like in Figure~\ref{fig:initialGame}.

\subsection{Gameplay}
\label{subsec:gameplay}
A match consists of nine rounds in which each player draws one card and potentially captures some of the cards on the table;
	accordingly, a round consists of four players' turns and a match consists of 36 turns.
The eldest hand plays first, then it is the turn of the dealer's teammate and so on counterclockwise.
At each turn the player must play a card from his hand. The chosen card can either be placed on the table or capture one or more cards from the table.
A capture is made by matching a card in the player's hand to a card of the same value on the table, or if that is not possible, by matching a card in the player's hand to the sum of the values of two or more cards on the table.
In both cases, both the card from the player's hand and the captured card(s) are removed and placed face down in a pile in front of the player.
Teammates usually share the same pile that is placed in front of one of them.
These cards are now out of play until the end of the match when scores are calculated.
For example, in Figure~\ref{fig:initialGame}, the player may choose to place on the table the $A\spadesuit$ or capture the $3\spadesuit$ with $3\diamondsuit$ or capture the $3\spadesuit$ and $5\diamondsuit$ with the $J\diamondsuit$.
Note that it is not legal to place on the table a card that has the ability to capture:
in Figure~\ref{fig:initialGame}, the player cannot place on the table the $Q\heartsuit$, because it can capture the $Q\spadesuit$.
In case the played card may capture either one or more cards, the player is forced to capture only the single card.
For example, in Figure~\ref{fig:initialGame}, the $Q\heartsuit$ cannot capture the $4\clubsuit$ and $5\diamondsuit$ because it is forced to capture the $Q\spadesuit$.
When the played card may capture multiple cards with different combinations, the player selects the combination she prefers.
When a capture removes all the cards on the table, the player has completed a \emph{scopa}, gaining one point, 
	and the card from the player's hand is placed face up under the player's pile
in order to take it into account when the scores are calculated; note that, at the last turn, \textit{scopa} is not allowed.
This move is called \textit{scopa} (in Italian it means \emph{broom}) because all the cards in the table are swept by the played card.
At the end of the match, when all the nine rounds have been completed, 
	all cards that are still on the table go to the last player who did a capturing move.

\subsection{Scoring}
\label{subsec:scoring}
When the match ends, the score of each team is calculated and added to the team overall game score.
There are five ways to gain points:
	(i) \emph{Scopa}, one point is awarded for each scopa;
	(ii) \emph{Cards}, the team who captured the largest number of cards receives one point;
    (iii) \emph{Coins}, the team who captured the largest number of cards in the suit of coins gets one point;
	(iv) \emph{Settebello}, the team who captured the seven of coins gets one point;
	(v) \emph{Primiera}, the team who obtained the highest \emph{prime} gets one point. 
	The prime for each team is determined by selecting the team's best card 
	in each of the four suits, and summing those four cards' point values.
	Table~\ref{tab:primiera} shows the cards' values used in this calculation.
	If a team has no cards in a suit, the point is awarded to the opponents, even if they have a minor sum.

The four points awarded with \textit{Cards}, \textit{Coins}, \textit{Settebello}, and \textit{Primiera} are called \emph{deck points}.
Note that, for both \textit{Cards}, \textit{Coins}, and \textit{Primiera}, in case of a tie, the point is not assigned.
Table~\ref{tab:scores} shows an example of a round's scores calculation.
The hand team's primiera is calculated on the cards $7\diamondsuit$ $7\spadesuit$ $5\heartsuit$ $7\clubsuit$ ($21+21+15+21=78$), whereas the deck team's primiera is calculated on the cards $6\diamondsuit$ $6\spadesuit$ $7\heartsuit$ $A\clubsuit$ ($18+18+21+16=73$).

\begin{table}[t!]
\centering
\caption{Cards values for the calculation of primiera.} 
\label{tab:primiera}
\scriptsize
\begin{tabular}{|r||c|c|c|c|c|c|c|c|c|c|}
\hline
	\textbf{Rank} & 7 & 6 & $A$ & 5 & 4 & 3 & 2 & $J$ & $Q$ & $K$ \\ 
\hline
	\textbf{Value} & 21 & 18 & 16 & 15 & 14 & 13 & 12 & 10 & 10 & 10 \\ 
\hline 
\end{tabular}
\end{table}
\begin{table}
\footnotesize
\centering
\caption{Example of scores calculation. Table (a) shows the cards in the pile of each team at the end of a match. 
Table (b) shows the points of each team and the final scores.}
\label{tab:scores}

\begin{tabular}{|c||cccccccc|}
\hline
	& \multicolumn{8}{c|}{Team's pile} \\
\hline
	\multirow{3}{*}{\textbf{Hand team}} &
		$5\spadesuit$ & $5\heartsuit$ & $4\diamondsuit$ & $A\spadesuit$ & $3\clubsuit$ & $3\diamondsuit$ & $3\heartsuit$ & $7\clubsuit$  \\
	&	$Q\spadesuit$ & $Q\heartsuit$ & $Q\clubsuit$ & $7\diamondsuit$ & $5\diamondsuit$ & $2\heartsuit$ & $4\spadesuit$ & $4\heartsuit$ \\ 
	&	$J\clubsuit$ & $4\clubsuit$  & $K\clubsuit$ & $6\clubsuit$ & $7\spadesuit$ & $Q\diamondsuit$ &  & \\ 
\hline
	\multirow{2}{*}{\textbf{Deck team}} & 
		$J\diamondsuit$ & $J\heartsuit$ & $K\diamondsuit$ & $K\heartsuit$ & $6\diamondsuit$ & $6\spadesuit$ & $J\spadesuit$ & $2\spadesuit$ \\
	&	$3\spadesuit$ & $2\clubsuit$ & $A\heartsuit$ & $A\diamondsuit$ & $K\spadesuit$ & $A\clubsuit$ & $2\diamondsuit$ & $7\heartsuit$ \\
	&	$6\heartsuit$& $5\clubsuit$ & & & & & & \\
\hline 
\end{tabular}

~\\
\centerline{(a)}
~\\

\begin{tabular}{|c|cc|}
\hline
\multicolumn{3}{|c|}{Match scores} \\
\hline
& \textbf{Hand team} & \textbf{Deck team} \\
\hline
\textbf{Scopa} & 1 & 3 \\
\textbf{Cards} & 22 & 18 \\
\textbf{Coins} & 5 & 5 \\
\textbf{Settebello} & 1 & 0 \\
\textbf{Primiera} & 78 & 73 \\
\hline
\textbf{Scores} & 4 & 3 \\
\hline
\end{tabular}
~\\
\centerline{(b)}
~\\

\end{table}


\section{Rule-Based Players for Scopone}
\label{sec:rulebasedai}
We developed three rule-based artificial players 
	(\emph{Greedy}, \emph{Chitarrella-Saracino}, and \emph{Cicuti-Guardamagna})
	implementing strategies of increasing complexity taken from the most important Scopone strategy books \cite{cicuti1978segreti,saracino2011scopone}.
The first one implements the basic strategy taught to beginners and, at each turn, 
	tries to perform the best capture available or it plays the least valuable card if a capture is not available.
The second one encodes a union of the rules written by Chitarrella and Saracino as reported in~\cite{saracino2011scopone}, 
	therefore it implements an expert strategy.
The third one encodes the rules written by Cicuti and Guardamagna \cite{cicuti1978segreti}
	that extend \cite{saracino2011scopone} by introducing  additional rules to handle of sevens.

\subsection{Greedy Strategy}
\label{sec:greedyAI}
The greedy strategy implements  a beginner player using the basic rules of the game and 
  whenever possible it captures the most important cards available, otherwise it plays the least important ones.
We defined card importance using a modified version of the cards' values for the primiera (Section~\ref{sec:scopone}) which ensures that the greedy player will try 
(i) to achieve the highest possible primiera score and to capture the $7\diamondsuit$ which
alone gets one point for the settebello;
(ii) to capture as many coins cards as possible to get the coin point.
The greedy player has a specific strategy to gain scopa points that tries 
(i) to block the opponents from doing a scopa; 
(ii) to decide whether to force or not the scopa when it is possible.
To deal with the former objective,
	the greedy player gives a higher priority to the moves
	that do not leave on the table a combination of cards 
	which could be captured by a card that is still in play
	(but it is in the hand of another player which is \emph{unknown} to the greedy player).
With respect to the latter issue, 
	a scopa move is never forced since often it may be better to skip it
	in favor of the capture of some important cards (e.g., coin cards, primeria cards).
However, even if not explicitly forced, a scopa move is usually performed as a defensive move
	not to leave on a table a combination of cards that might allow a scopa by the opponent players.

\subsection{Chitarrella-Saracino Strategy}
\label{sec:CSAI}
The \emph{Chitarrella-Saracino} (CS) strategy aims to behave like an expert player. 
It implements the 44 rules of Chitarrella and the most important playing strategies from the 110 advices contained in the book by Saracino~\cite{saracino2011scopone},
that have been summarized in~\cite{bampi}. In the following, we summarize the main aspects of the strategy, the detailed description of all the rules with the pseudo-code is available in \cite{dipalma:2014}.

\medskip\noindent\textbf{Basic Rules.} 
Scopone players are not allowed to speak and players have only the information they can gather from the cards that the other players have drawn.
Accordingly, Chitarrella and Saracino \cite{saracino2011scopone} include basic strategies to deal with the selection of the card to draw 
  when little or no information is available. Examples of such rules include,
	(i) in general, when it is possible to capture some cards, one has to do it;
	(ii) if it is not possible, one has to play a \emph{double card}, i.e. a card of which one also holds another card of the same rank;
	(iii) if the opponent captures such double card, and the teammate has the other card of the same rank, 
		the player's teammate must play the double card in order to create the \textit{mulinello} on that card 
		(in fact, the player is the only one who can capture the card played by the teammate, 
		since the other two were captured by the opponent).

\medskip\noindent\textbf{Spariglio} (decoupling in English)
	is one of the most important aspects of Scopone and consists in playing a card 
	that matches to the sum of the values of two or more cards on the table.
For example, if we capture 3 and 2 with 5, then we did the spariglio $3+2=5$.
Each card rank appears four times in the deck therefore,
	at the beginning of a match, all the cards are \emph{coupled}, 
	in the sense that, by doing moves not involving the spariglio, each card is taken by its copy 
	until there are no more cards on the table at the end of the round.
This advantages the deck team, because they always play after the opponents and also simplifies the choice of which cards to play.
When a player does a spariglio move, the involved cards that were \emph{coupled}, or \emph{even}, will become \emph{decoupled}.
Conversely, the involved cards that were \emph{decoupled}, or \emph{odd}, will return \emph{coupled}.
Spariglio moves make the selection of the best move more complex as the number of available options dramatically increases and, for human players, it makes more difficult to remember what cards have been played and what cards remain \cite{saracino2011scopone}.
An example of rules by Chitarrella and Saracino~\cite{saracino2011scopone} for spariglio include:
the hand team should play the last decoupled card of the highest rank, 
	this will limit the capturing options of the dealer who play last in the round;

\medskip\noindent\textbf{Mulinello}
(eddy in English) is a play combination in 
	which two players continuously capture while their opponents are forced to draw cards without performing any capture.
Mulinello often happens at the beginning of the match, when the eldest hand can do a good capture on the four cards on the table.
For example, let us assume that the table is $1\diamondsuit$ $3\heartsuit$ $3\clubsuit$ $6\spadesuit$.
The eldest hand, who has another 3 in her hand, captures $1\diamondsuit$ $3\heartsuit$ $6\spadesuit$ 
	with $K\diamondsuit$, challenging the fourth 3 because it holds the other one.
The dealer's teammate, to avoid a scopa, plays $Q\clubsuit$ that the third hand captures with $Q\heartsuit$.
The dealer follows the teammate and plays $Q\spadesuit$ that is captured by the eldest hand with $Q\diamondsuit$, and so on; 
	mulinello often happens when the deck team does not have double or triple cards. 
An example of strategic rules for mulinello \cite{saracino2011scopone} is:
if it is possible to do the mulinello either on a low-rank or face card, then the hand team must always prefer to do it on the low-rank card.

\medskip\noindent\textbf{The play of sevens} is very important as 7s count both for the primiera and the settebello points.
	Accordingly, Scopone is mostly played around these cards that hence deserve special rules.
Examples of rules devoted to the sevens include:
(i) the dealer's teammate must always capture the 7 played either by the dealer or by the opponents;
(ii) when the dealer's teammate has the chance to capture a 7 on the table or do a spariglio involving it, e.g. $7+1=8$, it must always do the spariglio if it holds only one 7;
(iii) if the dealer's teammate does not hold any 7 and has the chance to capture a 7 on the table with a spariglio, e.g. $7+1=8$, it must always do it even if it decouples three cards.

\medskip\noindent\textbf{Player Modeling.} Some of the Chittarella and Saracino rules assume that the player has memorized the cards played so far
	and based on the action of the players can guess what cards another player may or may not have. For example, if a player did not do a scopa move, 
	we can fairly assume that the card required for the scopa was not in her hand; or if a player to capture a card did not use a coin seed, we can assume
	she does not have the corresponding coin card. 
Accordingly, we implemented a \emph{card guessing} module that memorizes all the cards played by each player and estimates 
	what card may and may not be available to each player. The estimates are updated every time a player performs a move.

\subsection{Cicuti-Guardamagna Strategy}
\label{sec:CGAI}
The book of Cicuti and Guardamagna~\cite{cicuti1978segreti} refined the strategy of Saracino~\cite{saracino2011scopone} by introducing 
	advanced rules regarding the play of sevens~\cite{bampi,cicuti1978segreti}.
Some of such additional rules include:
(i) 
the player, who holds three 7s with the settebello, must immediately play one of them;
(ii) the player, who holds two 7s without the settebello, must play the 7 as late as possible;
(iii)
the player, who holds the last two 7s with the settebello,
	if it belongs to the hand team, it must play the settebello while if it belongs to the deck team, it must play the other 7.

\section{Monte Carlo Tree Search for Scopone}
\label{sec:mcts_scopone}
We evaluated several players for Scopone using MCTS and ISMCTS with different configurations.
We started with the basic MCTS algorithm that needs full knowledge of the
current game state and thus knows all the cards in the players' hand. 
This cheating player provided an upper bound of the performance achievable using an MCTS approach.
Next, we focused on \emph{Information Set Monte Carlo Tree Search} that deals with imperfect information and
implements a fair player.

\subsection{Basic Monte Carlo Tree Search}
\label{sec:MCTSbasic}
In the basic version of the MCTS algorithm for \textit{Scopone}, each node represents a single state of the game and memorizes: (i) the incoming move, (ii) the visits count, (iii) the total rewards, (iv) the parent node, and (iv) the child nodes.
Each state memorizes the list of cards on the table, in the players hands, and the cards captured by each team.
It also memorizes the cards that caused a scopa, the last player to move, and the last player who did a capturing move.
Each state also records all the moves previously done by each player along with the state of the table before each move. 
%
A move comprises the played card and the list of captured cards, which can be empty.
Each card is represented by its rank and suit.
Every state can return the list of available moves and, in case of a terminal state, the score of each team.
The selection step of MCTS uses UCT (Eq. \ref{eq:uct}).
In the expansion step, the move to expand the node is chosen randomly from the available ones.
For the simulation step the game is played randomly until a terminal state.
The backpropagated reward is a vector containing the score of each team.
During the selection step, UCT considers only the team's score of the player acting in that node.

\subsection{Basic Information Set Monte Carlo Tree Search}
\label{sec:ISMCTSbasic}
In the basic version of the ISMCTS player for Scopone, each node represents an information set from the root player's point of view. The only additional information required to each node, with respect to MCTS, is the availability count.
The ISMCTS algorithm applies the same steps of MCTS, but it uses the ISUCT (Eq.~\ref{eq:uct-ismcts}).
It also creates a determinization of the root state at each iteration by randomizing 
	the cards held by the other players.


\subsection{Reward Function}
\label{sec:reward}
When a terminal state is reached (a match is ended), 
  a reward function is applied to compute a reward that is then backpropagated to the nodes
  visited during the selection and expansion steps.
In this study, we considered four reward functions: 
	\emph{Raw Scores} (RS) returns the score of each team as reward;
	\emph{Scores Difference} (SD), for each team returns the difference between the score of that team and the opponent team 
	(for example, if the final scores of a game were 4 and 1, the reward for the first team would be $3$ and $-3$ for the other one);
	\emph{Win or Loss} (WL) returns a reward of $1$ to the winning team, $-1$ to the other one, and $0$ in case of a tie.
	\emph{Positive Win or Loss} (PWL) returns a reward of $1$ for the winning team and a reward of $0$ for the losing team; 
	in case of a tie, both teams receive a reward of $0.5$.
Note that players using RS and SD will try to achieve the maximum score even if they are going to lose the game, following a more human-like behavior.
PWL is similar to WL, but we want to exploit the fact that, for rewards between $[0,1]$, the optimal UCT constant is known \cite{DBLP:journals/tciaig/BrownePWLCRTPSC12}.

\subsection{Simulation Strategy}
\label{sec:simulationStrategies}
The simulation strategy is responsible to play the moves from a given state until a terminal state. 
In this study, we considered four simulation strategies.
\emph{Random Simulation} (RS) simulates opponents' moves by selecting random actions from the ones available in the current state. 
%
We also introduced three heuristic strategies which according to previous studies might increase the performance of MCTS approaches \cite{magic}.
\emph{Card Random Simulation} (CRS) plays a card at random, but applies the Greedy strategy to decide which capturing move to do in case there are more than one;
\emph{Greedy Simulation} (GS) applies the Greedy strategy (Section~\ref{sec:greedyAI});
\emph{Epsilon-Greedy Simulation} (EGS), at each turn, plays at random with probability $\epsilon$, otherwise it plays the move selected by the Greedy strategy.

\subsection{Determinization with the Cards Guessing System}
\label{sec:determinCGS}
To reduce the complexity of the ISMCTS tree, we can decrease the number of states within an information set
  by removing the states that are less likely to happen so that the search will be focused only in the most likely states.
Accordingly, in the determinization step of ISMCTS, we integrated the cards guessing system that we designed for the Chitarrella-Saracino strategy (Section~\ref{sec:rulebasedai}). Therefore, at each iteration of ISMCTS, in the generated determinization each player holds the cards suggested by the cards guessing system.

\section{Experimental Results}
\label{sec:experiments}
We performed a set of experiments to evaluate all the artificial players we developed for \Scopone. 
At first, we evaluated the game bias towards deck and hand teams using players adopting a random strategy.
Next, we performed a set of experiments to determine the best player for each type of artificial intelligence (rule-based, MCTS, and ISMCTS).
At the end, we performed a tournament to rank the players selected in the previous steps
	and an experiment involving human players.

\subsection{Experimental Setup}
\label{sec:experimentalSetup}
To provide consistency among different sets of experiments and to reduce the variance of the results, 
  we randomly generated 1000 initial decks (that is, initial game states) using a uniform distribution and used these 1000 decks
  in all the experiments presented in this paper.
Moreover, 
   when using MCTS and ISMCTS, we played each match 10 times to further reduce the variance due to the stochastic nature of the algorithms.
In each experiment, first we assigned one artificial intelligence to the hand team and the other one to the deck team,
	next we repeated the same experiment switching the hand/deck roles, 
	finally we compared the winning rate for each team and the percentage of ties.
For each experiment, we report the winning rate for each team and the $95\%$ confidence interval in square brackets.

\subsection{Random Players}
\label{sec:ExpRandom}
As the very first step, we used players following a random strategy for both the hand and deck teams
  to evaluate how much the game is biased towards one of the roles.
The results show that the deck team has an advantage over the hand team, 
   winning $45.7\% [45.3 - 46.1]$ of the matches, while the hand team wins $41.7\% [41.3 - 42.1]$
   of the matches and $12.6\% [12.4 - 12.8]$ of the matches end with a tie.
Such advantage was known and already mentioned in the historical strategy books \cite{chitarrella2002regole,saracino2011scopone}, 
	but was not estimated quantitatively before.
Note however that the reported difference is not statistically significant for a 95\% confidence level (p-value is 0.071).

\begin{table}[h!]
\centering
\caption{Tournaments involving the three rule-based players. The hand team is listed in the left column while the deck team is listed at the top. The first table shows the percentage of wins of the hand team, the second table shows the percentage of losses of the hand team, and the third table shows the percentage of ties.
The corresponding 95\% confidence intervals are reported in square brackets.}

\footnotesize

\label{tab:RBAITournament}
\begin{scriptsize}
\begin{tabular}{|c|c|c|c|}
\hline
\multicolumn{4}{|c|}{\textbf{Winning rates of the hand team}} \\
\hline
$\nicefrac{\textbf{Hand Team}}{\textbf{Deck Team}}$& \textbf{Greedy} & \textbf{CS} & \textbf{CG} \\
\hline
\textbf{Greedy} & 
39.3\% &
36.5\% & 
37.7\% \\
& 
[38.9\% - 39.7\%] &
[36.1\% - 36.9\%] & 
[37.3\% - 38.1\%] \\
\hline
\textbf{CS} & 
46.7\% &
38.7\% &
37.5\% \\
	& 
[46.3\% - 47.1\%] &
[38.3\% - 39.1\%] &
[37.1\% - 37.9\%] \\
\hline
\textbf{CG} & 
43.7\% &
37.9\% &
37.3\% \\
& 
[43.3\% - 44.1\%] &
[37.5\% - 38.3\%] &
[36.9\% - 37.7\%] \\
%
%
%
\hline
\end{tabular}
\end{scriptsize}
~\\
\vspace*{0.1 cm}
(a)\\
\vspace*{0.3 cm}
\begin{scriptsize}

\begin{tabular}{|c|c|c|c|}
\hline
\multicolumn{4}{|c|}{\textbf{Losing rates of the hand team}} \\
\hline
$\nicefrac{\textbf{Hand Team}}{\textbf{Deck Team}}$& \textbf{Greedy} & \textbf{CS} & \textbf{CG} \\
\hline
\textbf{Greedy} & 
43.4\% &
48.8\% &
48.2\% \\
& 
[43.0\% - 43.8\%] &
[48.4\% - 49.2\%] &
[47.8\% - 48.6\%] \\
\hline
\textbf{CS} &
40.1\%  &
46.6\%  &
47.6\%  \\
&
[39.7\% - 40.5\%] &
 [46.2\% - 47.0\%] &
[47.2\% - 48.0\%] \\
\hline
\textbf{CG} & 
42.0\%		&
48.2\%		&
49.1\%		\\
& 
[41.6\% - 42.4\%] &
[47.8\% - 48.6\%] &
[48.7\% - 49.5\%] \\
\hline
\end{tabular}

\end{scriptsize}
~\\
\vspace*{0.1 cm}
(b)\\
\vspace*{0.3 cm}
\begin{scriptsize}

\begin{tabular}{|c|c|c|c|}
\hline
\multicolumn{4}{|c|}{\textbf{Tying rates}} \\
\hline
$\nicefrac{\textbf{Hand Team}}{\textbf{Deck Team}}$& \textbf{Greedy} & \textbf{CS} & \textbf{CG} \\
\hline
\textbf{Greedy} & 
17.3\% & 
14.7\% & 
14.1\% \\
& 
[17.0\% - 17.6\%] & 
[14.5\% - 14.9\%] & 
[13.9\% - 14.3\%] \\
\hline
\textbf{CS} & 
13.2\% & 
14.7\% & 
14.9\% \\
& 
[13.0\% - 13.4\%] & 
[14.5\% - 14.9\%] & 
[14.7\% - 15.1\%] \\
\hline
\textbf{CG} &
14.3\%	& 
13.9\% &
13.6\% \\
 &
[14.1\% - 14.5\%] & 
[13.7\% - 14.1\%] &
[13.4\% - 13.8\%] \\
\hline
\end{tabular}
\end{scriptsize}
~\\
\vspace*{0.1 cm}
(c)\\
\end{table}


\begin{table}
\caption{Percentage of wins, losses, and ties for each artificial intelligence involved in the final tournament.
The corresponding 95\% confidence intervals are reported in square brackets.
\label{tab:RBAIScoreboard}}
\centering
\begin{scriptsize}
\begin{tabular}{|c||c|c|c|}
\hline
\textbf{AI} & \textbf{Wins} & \textbf{Losses} & \textbf{Ties} \\
\hline
\hline
\textbf{CS} & 
44.4\% &
41.2\% &
14.3\% \\
& 
[44.1\% - 44.7\%] &
[40.9\% - 41.5\%] &
[14.2\% - 14.5\%] \\
\hline
\textbf{CG} & 
44.0\%			& 
42.0\%			& 
14.1\%			\\
					& 
[43.7\% - 44.3\%]	& 
[41.7\% - 42.3\%]	& 
[13.9\% - 14.2\%]	\\
\hline
\textbf{Greedy} & 
39.8\% & 
45.0\% & 
15.2\%	\\
& 
[39.6\% - 40.1\%] & 
[44.7\% - 45.3\%] & 
15.0\% - 15.3\%] \\
\hline
\end{tabular}
\end{scriptsize}
\end{table}

\subsection{Rule-Based Artificial Intelligence}
\label{sec:ExpRuleBased}
The second set of experiments was aimed at selecting the best rule-based AI among the ones considered in this study (Section~\ref{sec:rulebasedai}).
For this purpose, we performed a tournament between all three rule-based AIs we developed: 
  \emph{Greedy}, \emph{Chitarrella-Saracino} (CS), and \emph{Cicuti-Guardamagna} (CG).
The greedy strategy behaves like the typical human player who has been just introduced to the game and 
  basically tries to capture as many valuable cards as possible while trying to block scopa moves. 
CS and CG represent expert players of increasing complexity (CG extends CS with additional rules for the play of sevens).
Table~\ref{tab:RBAITournament} shows the results of the tournament involving 1000 matches played for each showdown (i.e., 1000 matches for each table position); 
Table~\ref{tab:RBAIScoreboard} shows the final scoreboard.
Unsurprisingly, the Greedy strategy performs worst, but it turns out to be stronger than one might expect.
In fact, the Greedy strategy wins the $39.83\%$ of the games and there is only a difference of about $4\%$ of wins from the other two AIs.
Probably, the scopa prevention and the playing of the best move considering only the importance of the cards are sufficient to obtain a good strategy for Scopone.
This is also why Scopone is more popular than other card games among young players: in fact, as our results show, 
  a beginner can easily win several games even against more expert players just by applying the basic rules and a rather elementary defensive strategy against opponents' scopa.
Furthermore, we note that when Greedy plays against a Greedy opponent, it ties more matches than against CS and CG
	(Table~\ref{tab:RBAITournament}c): $17.3\%$ versus $14.7$ and $13.6\%$ (although the difference is not statistically significant).
This may be explained by the fact that the Greedy strategy seeks to capture as much and as best cards as possible, and this leads 
to more frequent ties since the points are equally distributed.

The results also show that the CS and CG strategies have almost the same playing strength, they win about the $44\%$ of the games.
This was kind of unexpected since CG is supposed to be an improvement over the CS strategy as it adds advanced rules for the play of sevens.
These results suggest once more that special rules for the play of sevens might not be needed since CS somehow already handles them.
Thus, we selected CS as the best rule-based AI to be used in the final tournament against MCTS and ISMCTS approaches.

\subsection{Monte Carlo Tree Search}
\label{sec:ExpMCTS}
The next set of experiments was aimed at selecting the best setup for the MCTS player that is
	(i) the UCT constant,
	(ii) the reward function, and
	(iii) the simulation strategy.
In all the experiments, MCTS played 1000 matches as the hand team and then 1000 matches as the deck team. As usual, 
95\% confidence intervals are reported in square brackets.

\begin{figure*}[t]
\centering
\begin{tabular}{cc}
\parbox{.5\textwidth}{
\includegraphics[width=\plotwidth]{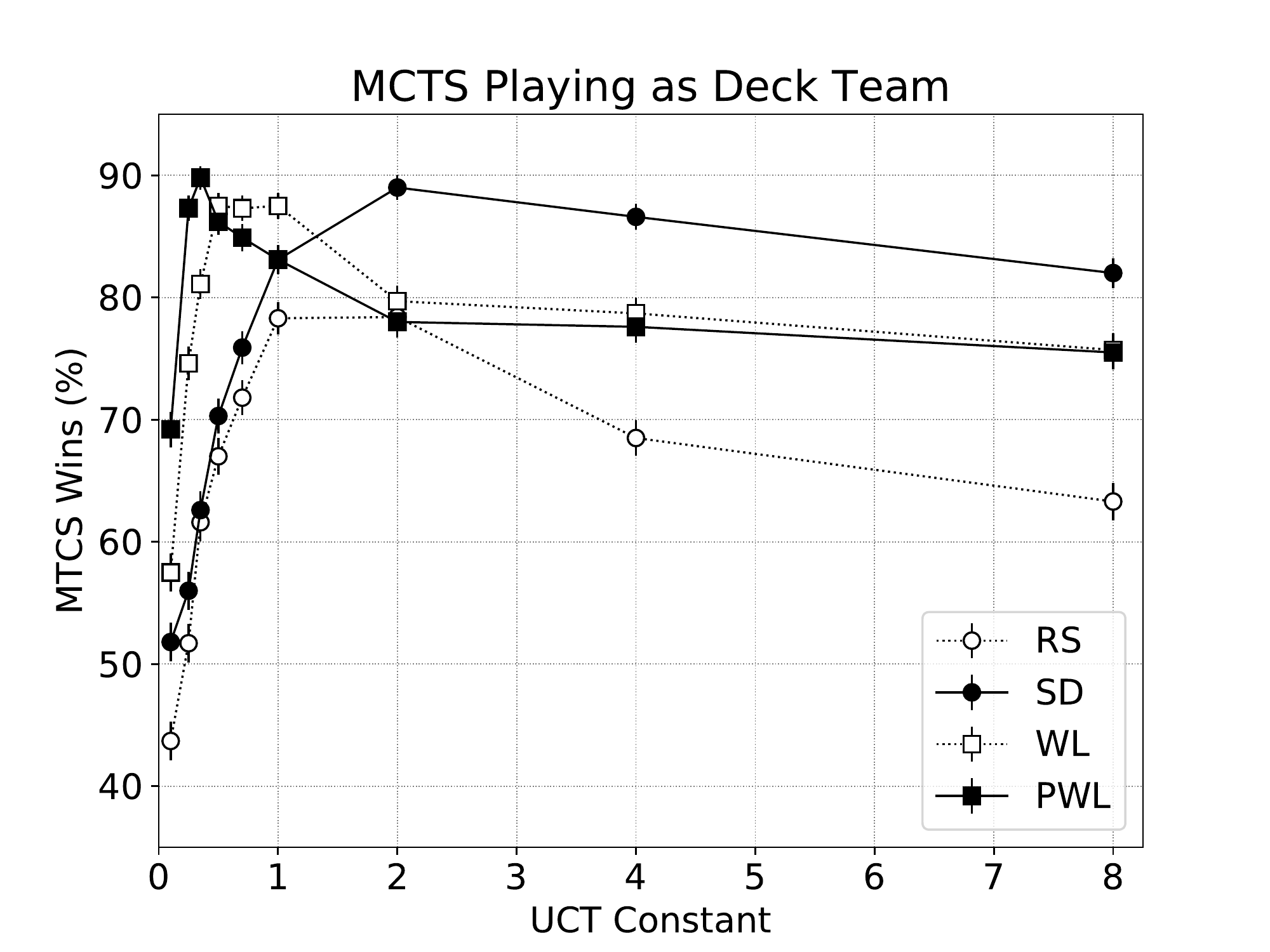}

\centerline{(a)}
}
&
\parbox{.5\textwidth}{
\includegraphics[width=\plotwidth]{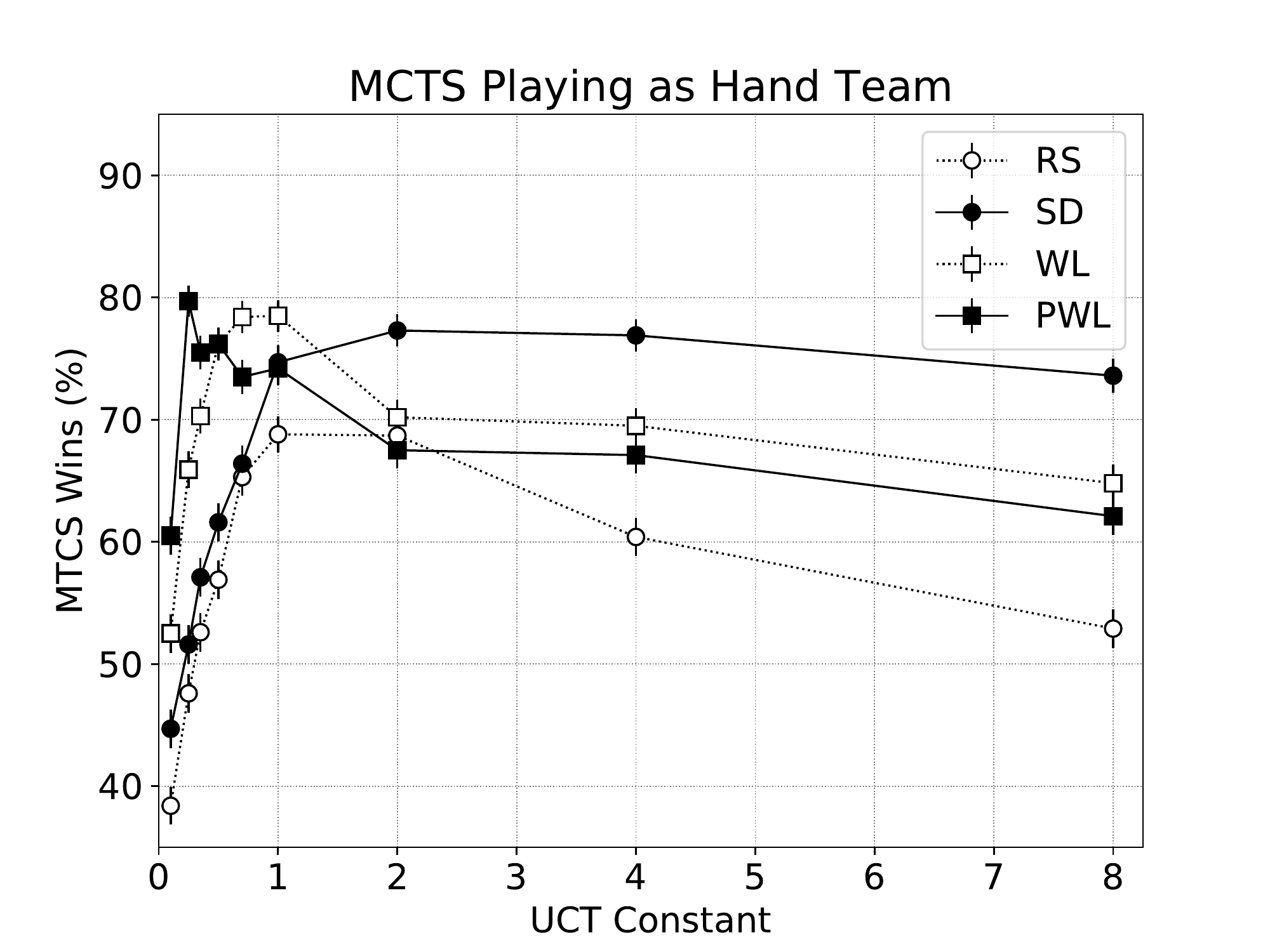}

\centerline{(b)}
}\\
\end{tabular}
\caption{Winning rate for MCTS using WL, PWL, RS, and SD with 1000 iterations and different values of UCT constant when playing against the Greedy strategy as (a) the deck team and (b) the hand team; bars report the standard error.}
\label{fig:MCTSUCT}
\end{figure*}

\medskip\noindent\textbf{Upper Confidence Bounds for Trees Constant.}
At first, we performed a set of experiments to select the best UCT constant for all the reward functions considered in this work namely, 
\emph{Win or Loss} (WL), \emph{Positive Win or Loss} (PWL), \emph{Scores Difference} (SD), and \textit{Raw Score} (RS).
We compared the performance of MCTS using the four reward functions, 1000 iterations,\footnote{We selected 1000 iterations because a set of preliminary experiments showed that is when MCTS performance begins to stabilize.} and different values of the UCT constant (0.1, 0.25, 0.35, 0.5, 0.70, 1.0, 2.0, 4.0, and 8.0) against
the Greedy strategy.
Figure~\ref{fig:MCTSUCT} reports the winning rate as a function of the UCT constant when MCTS plays as (a) the deck team and as (b) the hand team; bars report standard error values. 
The plots are similar apart from the higher winning rate of the deck teams, which was expected
	as we already showed that the game has a bias in favor of the deck team. 
\textit{Raw Score} (RS) is the worst performing reward function and thus we did not include it in the following experiments.
When playing as the deck team (Figure~\ref{fig:MCTSUCT}a),
	PWL achieved its best performance at 0.35 with a winning rate of $89.8\% [87.9-91.7]$
		and the difference with the performance for 0.25 and 0.5 is statistically significant; 
	WL between 0.5 and 1.0 with a winning rate of $87.5\% [85.5-89.5]$ for 0.7;
		and SD at 2.0 with $89.0\% [87.1-90.9]$.
The results for the hand team (Figure~\ref{fig:MCTSUCT}b) are similar apart that
	there is no peak for PWL at 0.35 ($75.5\% [72.8-78.2]$) which has a similar performance for 0.50 ($76.2\% [73.6-78.8]$);
	PWL still performs similarly when the UCT constant is set to 0.7 or 1.0 while it drops for 0.5.
 
At the end, for the next experiment, we selected 
	(i) \emph{Scores Difference} (SD) with an UCT constant of 2;
	(ii) \emph{Win or Loss} (WL) with an UCT constant of 0.7;
	(iii) \emph{Positive Win or Loss} (PWL) with an UCT constant of 0.35.

\begin{figure*}[t]

\centering
\begin{tabular}{cc}
\parbox{.5\textwidth}{
\includegraphics[width=\plotwidth]{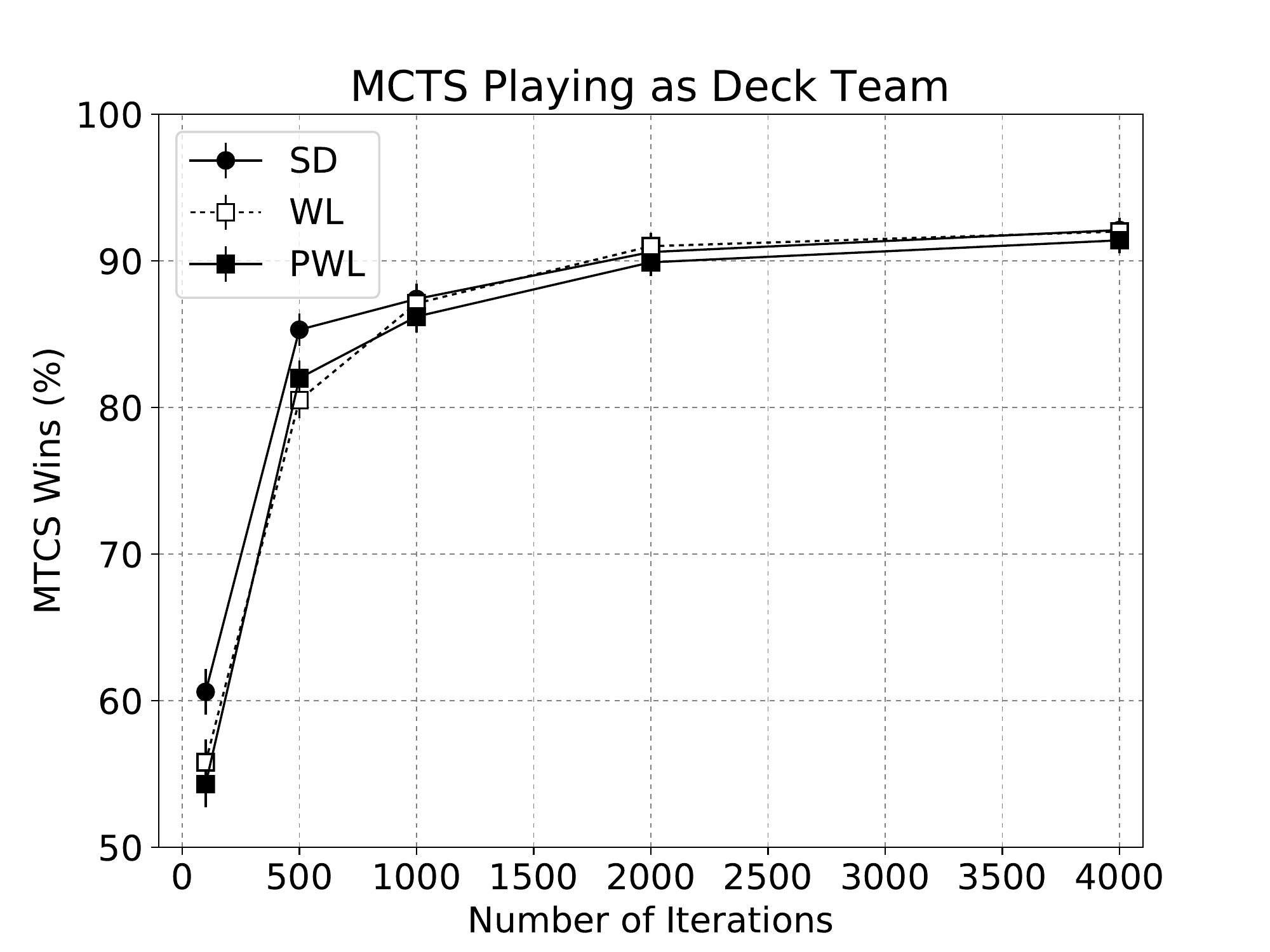}

\centerline{(a)}
} & 
\parbox{.5\textwidth}{
\includegraphics[width=\plotwidth]{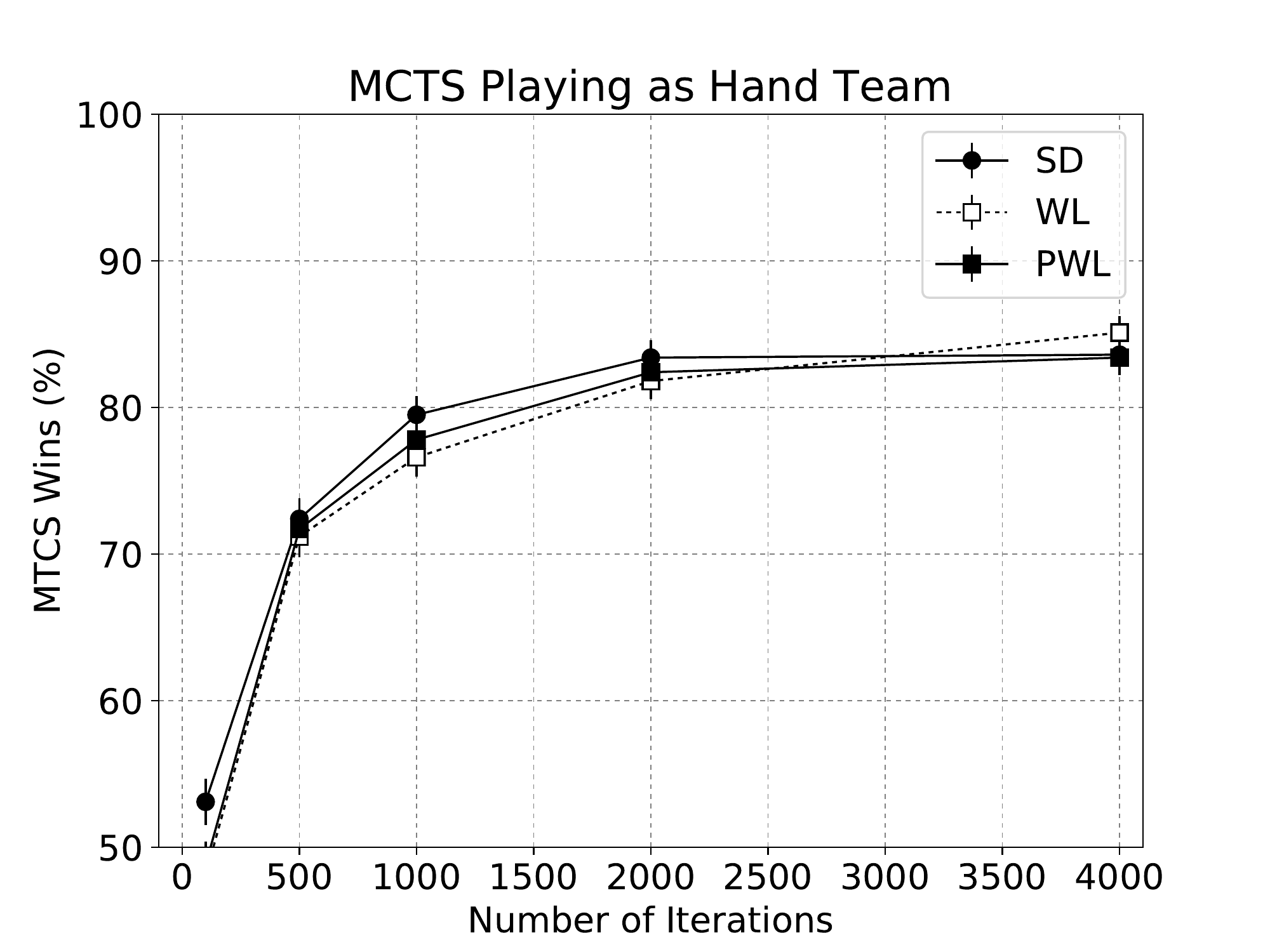}

\centerline{(b)}
}\\
\end{tabular}

\caption{Comparison of reward functions for the Monte Carlo Tree Search. Winning rate as a function of the number of iterations when MCTS 
  plays against the Greedy strategy as (a) the deck team and as (b) the hand team; bars report the standard error.}

\label{fig:MCTSrewards}

\end{figure*}

\medskip\noindent\textbf{Reward Functions.}
Next, we compared the performance of MCTS using the three reward functions using the UCT constants 
	previously selected.
Figure~\ref{fig:MCTSrewards} reports the winning rate of the three versions of MCTS 
	as a function of the number of iterations when MCTS plays as (a) the deck team and (b) the hand team.
The plots show that, as the number of iterations increases the three reward functions perform similarly 
	with no statistically significant difference beyond 1000 iterations.
When playing as the deck team, the three approaches reach 
		an average winning rate of $91.8\% [90.1-93.5]$ with 4000 iterations and $93.8\% [92.3-95.3]$ with 32000 iterations (not reported in the figure);
	when playing as the hand team, they reach an average winning rate of $84.0\% [81.8-86.3]$ with 4000 iterations and $86.2\% [84.0-88.3]$ with 32000 iterations.
With 500 iterations, when playing as the deck team, SD reaches a winning rate of $85.3\% [83.11-87.49]$ performing 
	significantly better than WL ($80.5\%	[78.0-83.0]$) with a p-value of 0.004;
	the difference between SD and PWL ($82.0\% [79.6-84.4]$) is much smaller and its statistical significance is borderline with a p-value of 0.046, 
	while the difference in performance between WL and PWL is not statistically significant.
With 100 iterations, SD still performs significantly better than PWL and WL.
Overall, the results suggest that, given enough iterations, 
	the three reward functions perform similarly and that, with fewer iterations, 
	SD might perform slightly better than PWL and WL;
	accordingly, we selected SD for the next experiments.
Finally, we note that MCTS outperforms CS when facing greedy players (see Section~\ref{sec:ExpRuleBased}) which was expected as MCTS implements a cheating player that has a complete knowledge of the game state.	

\medskip\noindent\textbf{Simulation Strategies.}
MCTS requires the estimation of the state's value of a leaf node.
The simulation strategy is responsible to play a game from a given state until the end and obtain an approximation of the state's value.
Previous studies \cite{magic} suggested that using heuristics in the simulation step can increase the performance of the algorithm.
Therefore, we performed a set of experiments to compare four simulation strategies:
\emph{Random Simulation} (RS),
\emph{Card Random Simulation} (CRS),
\emph{Greedy Simulation} (GS), and \emph{Epsilon-Greedy Simulation} (EGS). 
In particular, we used the standard simulation strategy (RS) as the baseline to compare the improvement provided by the other three heuristics.

\begin{figure}
\centering
\includegraphics[width=\plotwidth]{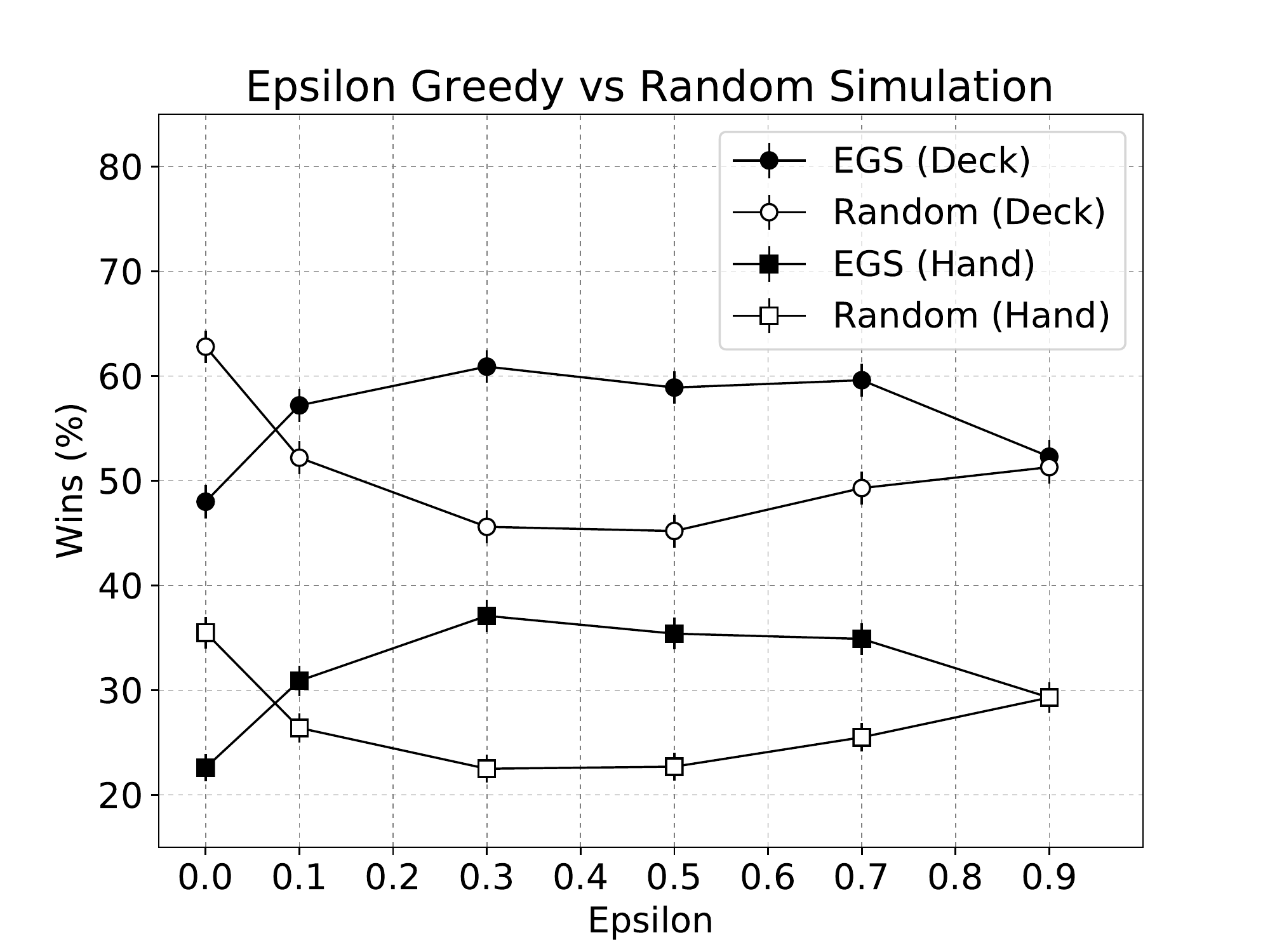}

\caption{Winning rate of MCTS using SD with a UCT constant of 2 as a function of the $\epsilon$ value used by the Epsilon-Greedy Simulation strategy when it plays against the same MCTS using random simulation. 
Bars report the standard error.
\label{fig:MCTSEGS}}
\centering

\includegraphics[width=\plotwidth]{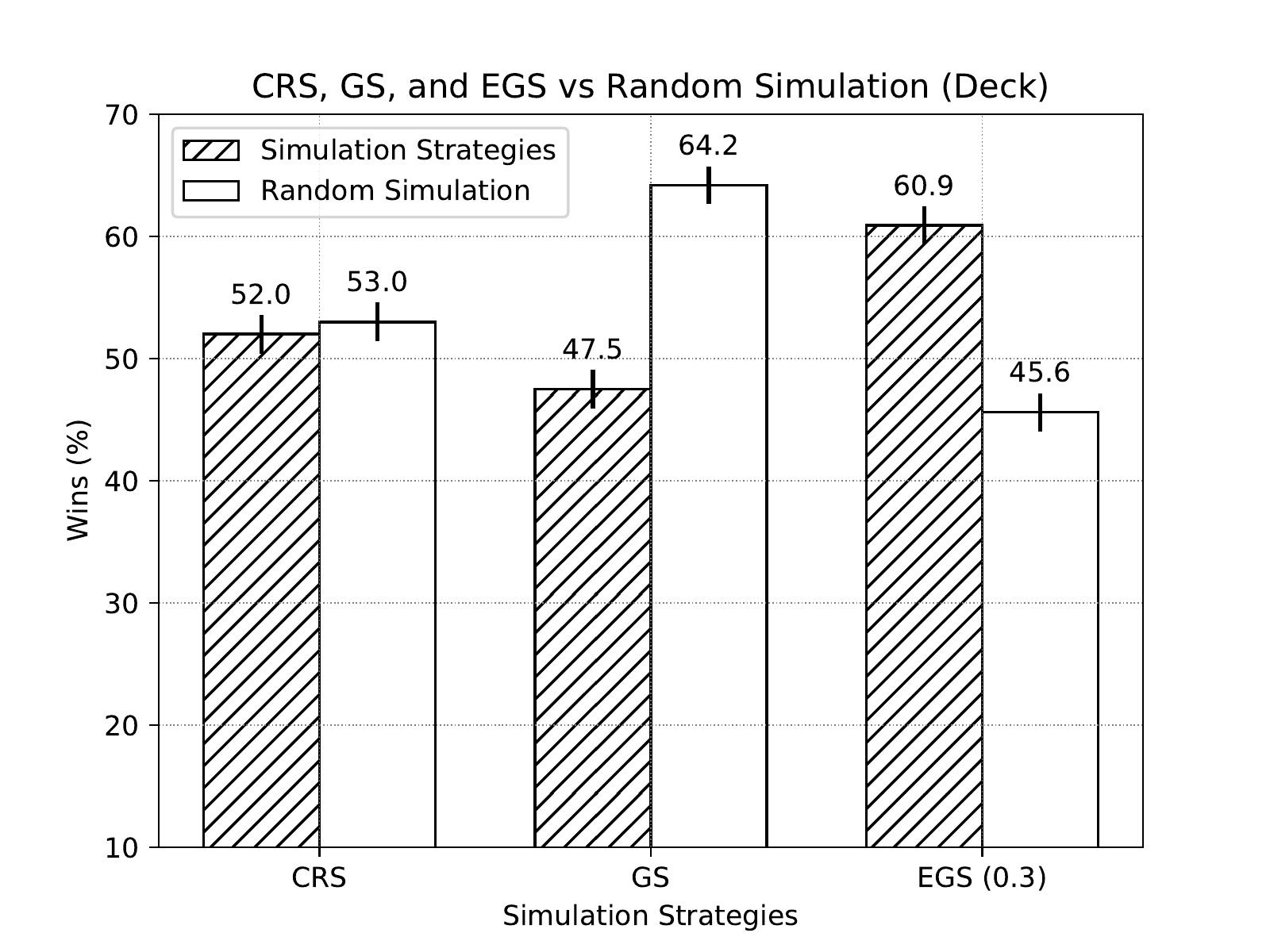}

\centerline{(a)}

\includegraphics[width=\plotwidth]{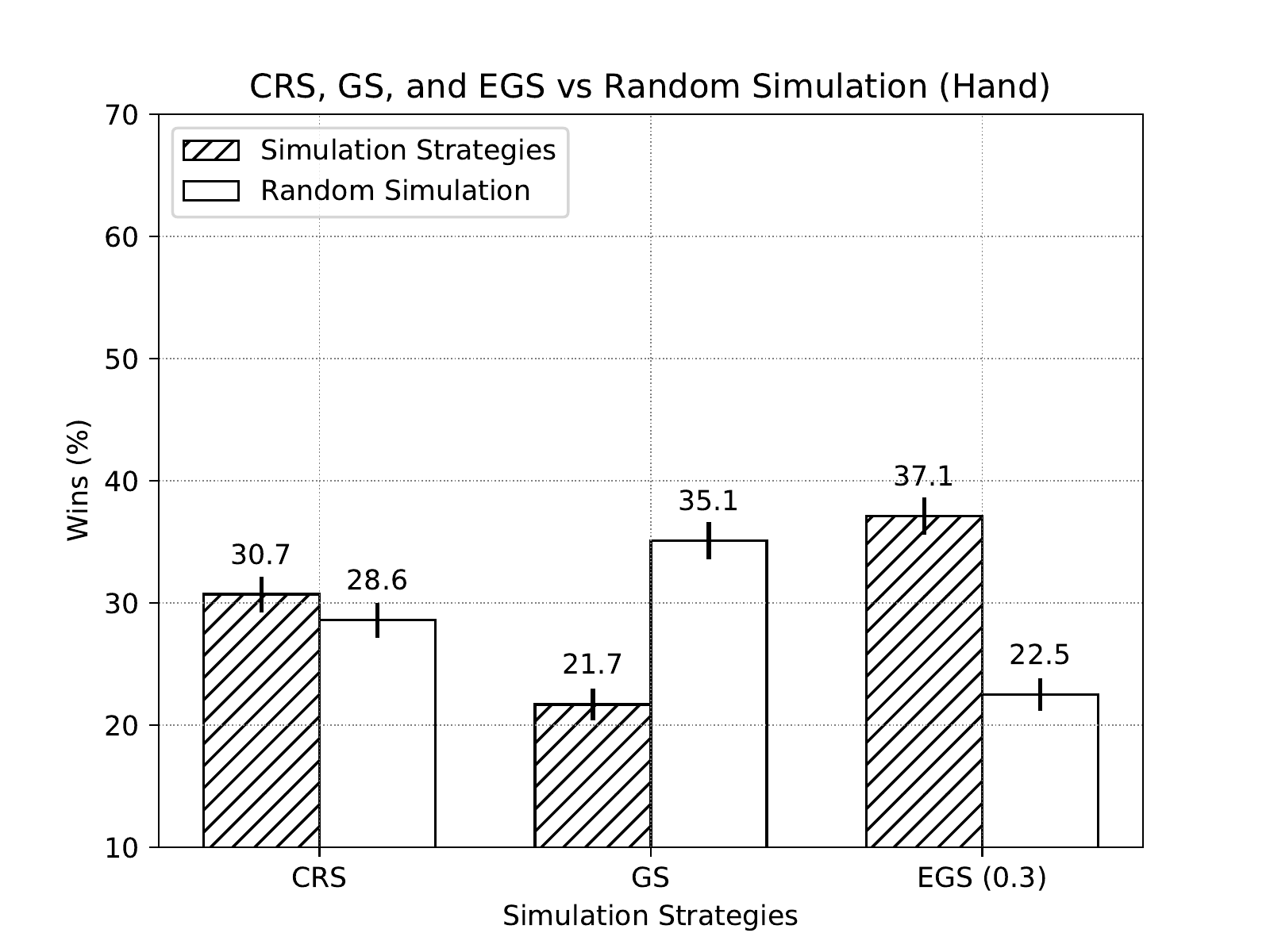}

\centerline{(b)}

\caption{Winning rate of MCTS using CRS, GS and EGS with an epsilon of 0.3 compared to MCTS using random simulation when playing as (a) the deck and (b) the hand team. Bars report the standard error.
\label{fig:MCTSSimulationStrategy}}
\end{figure}

At first, we performed an experiment to select the best value of $\epsilon$ for EGS. 
We matched MCTS using SD and Epsilon-Greedy simulation (EGS) with different values of $\epsilon$ (0.0, 0.1, 0.3, 0.5, 0.7, and 0.9) against the same MCTS using Random simulation (RS).
Note that, in contrast with the previous experiments, 
	in this case we did not use the Greedy strategy as the baseline because, when using EGS simulation, we would have implicitly provided a model of the opponent to MCTS
	and this would have biased our evaluation. 
Figure~\ref{fig:MCTSEGS} compares the performance of MCTS using EGS (solid markers) with that of MCTS using random simulation (white markers)
	when playing as the deck team (circle markers) and the hand team (square markers).
EGS outperforms plain random simulation for values of $\epsilon$ between 0.3 and 0.7 
	both when playing as the deck team (upper solid line) and as the hand team (lower solid line);
	when $\epsilon$ is 0.9, EGS selects a random action 90\% of the times and thus its behavior becomes almost identical to RS;
	when $\epsilon$ is 0.0, EGS always uses the deterministic Greedy player for simulations and, as the results show, this dramatically harms EGS performance.
EGS reaches its best performance at 0.3 both when playing for the deck team ($60.9\% [57.9-63.9]$) and for the hand team ($37.1 [34.1-40.1]$)
	and therefore we selected it for the next comparison.
Note however that, EGS performance for 0.3 is not statistically significantly different than what achieved with an $\epsilon$ of 0.5 ($58.9\% [55.9-62.0]$ and $35.4\%	[32.4-38.4]$) and 0.7 ($59.6\% [56.6-62.6]$ and $34.9\%	[32.0-37.9]$).
Finally, we note that the winning rates of the hand players (Figure~\ref{fig:MCTSEGS}) are much lower
	than the ones reported for random and rule-based players (Section~\ref{sec:ExpRandom} and Table~\ref{tab:RBAITournament})
	suggesting that having an advanced deck player increases the existing game bias toward this role.

Figure~\ref{fig:MCTSSimulationStrategy} compares the performance of random simulation against 
	\emph{Card Random Simulation} (CRS),
	\emph{Greedy Simulation} (GS), and \emph{Epsilon-Greedy Simulation} (EGS) with $\epsilon=0.3$ when playing as (a) the deck and (b) the hand team.
All the experiments employed the SD reward function with a UCT constant of 2.
EGS outperforms plain random strategy both when playing as the hand and as the deck team. Greedy strategy always performs worse than random simulation, confirming the results in Figure~\ref{fig:MCTSEGS} when $\epsilon$ was 0; CRS includes some knowledge about the game, but probably it is not sufficient to give some advantages, in fact it is almost equivalent to RS.
For these reasons, we fixed the EGS strategy with $\epsilon=0.3$ for the final tournament. 


\subsection{Information Set Monte Carlo Tree Search}
\label{sec:ExpISMCTS}
We repeated the same experiments using ISMCTS.
In this case, ISMCTS does not know the other players' hands (it does not cheat like MCTS) and 
	it bases its decisions solely on a tree search, where each node is an information set representing all 
	the game states compatible with the information available to the ISMCTS player.

\begin{figure*}

\begin{tabular}{cc}
\parbox{\columnwidth}{
\centering
\includegraphics[width=\plotwidth]{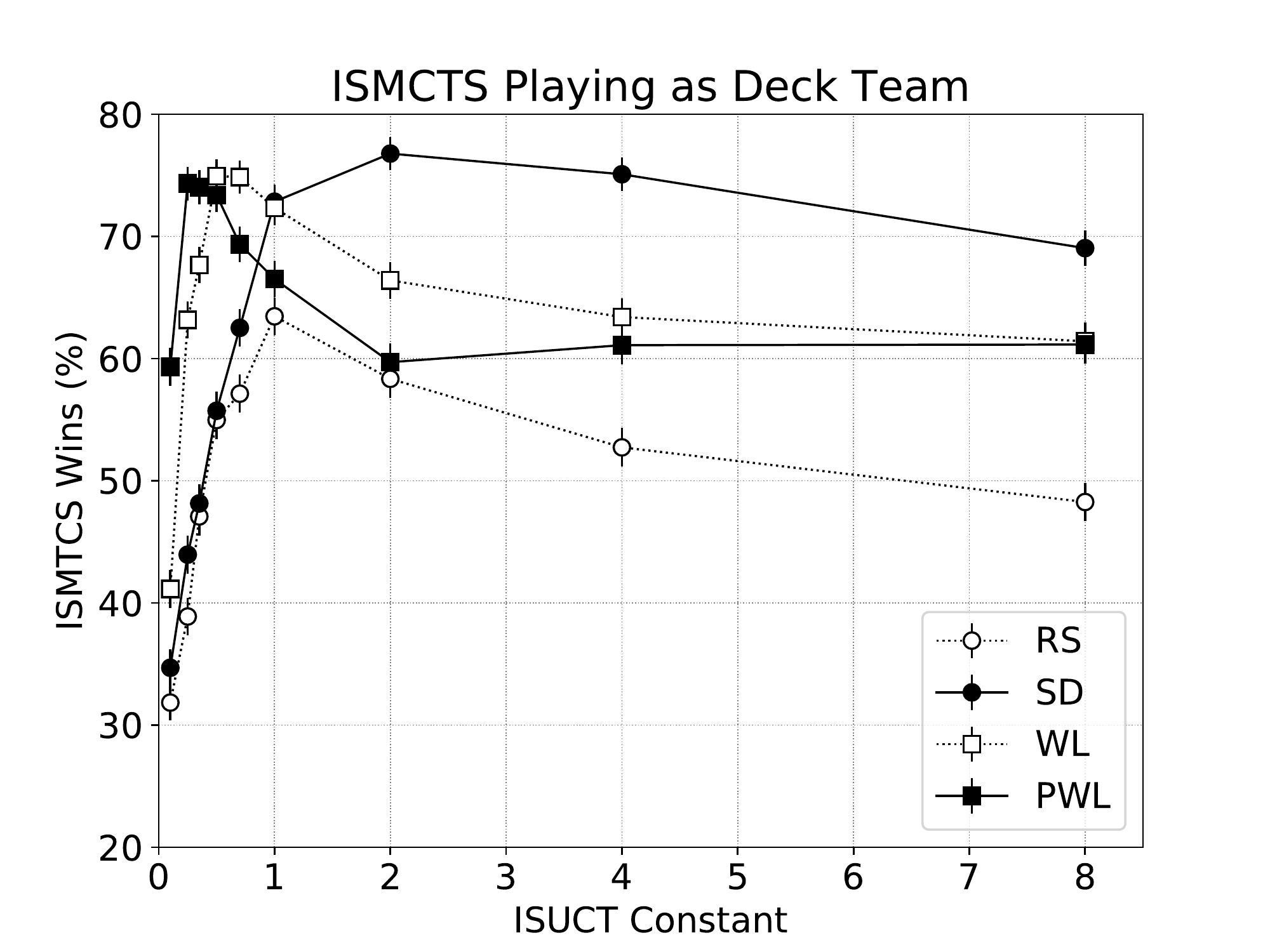}

\centerline{(a)}
} & 
\parbox{\columnwidth}{
\centering

\includegraphics[width=\plotwidth]{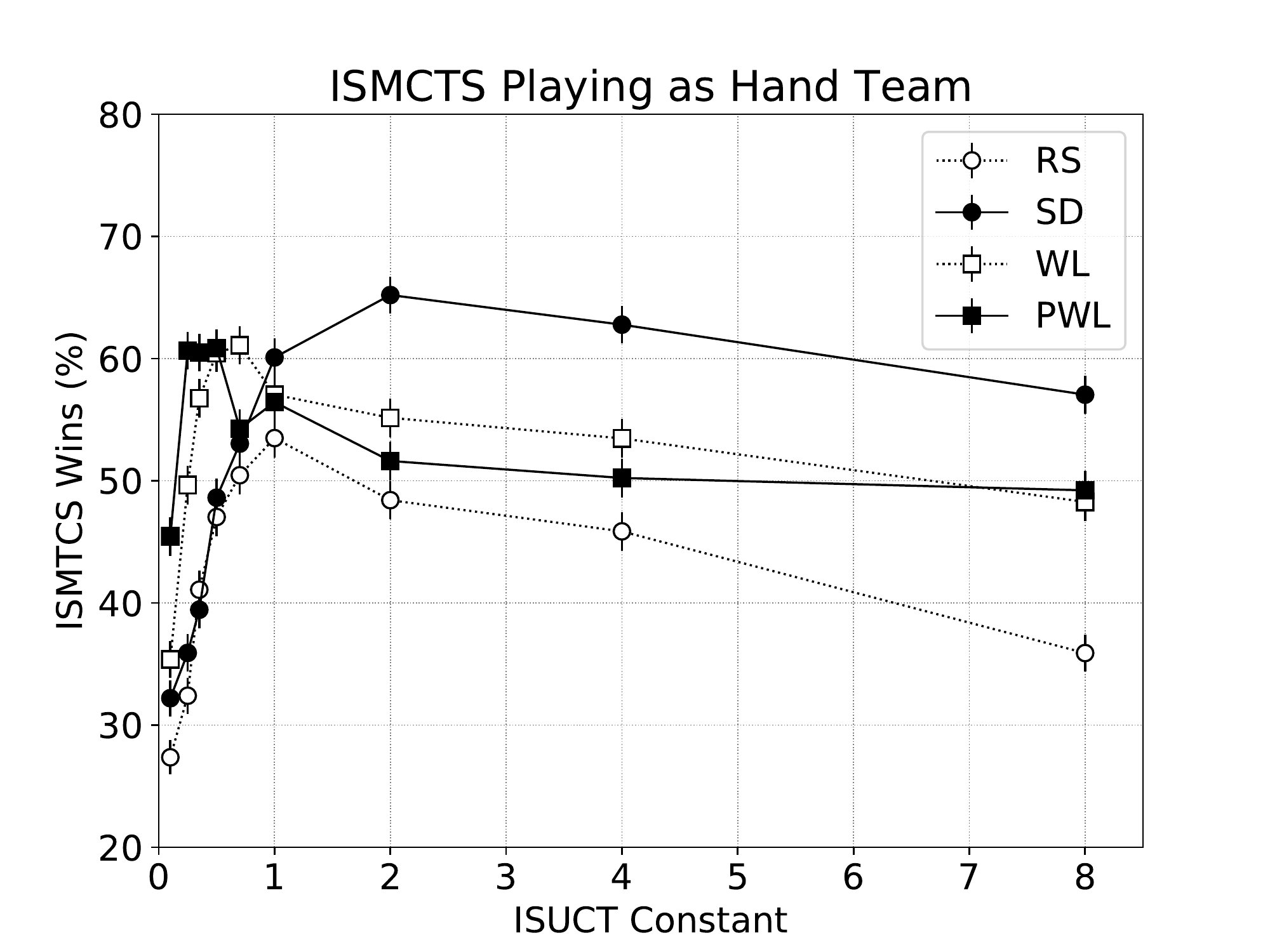}

\centerline{(b)}
}\\
\end{tabular}

\caption{Winning rate for ISMCTS using WL, PWL, RS, and SD with 4000 iterations and different values of ISUCT constant when playing against the Greedy strategy as (a) the deck team and (b) the hand team; bars report the standard error.}
\label{fig:ISMCTSUCT}
\end{figure*}

\medskip\noindent\textbf{Upper Confidence Bounds for Trees Constant.}
Figure~\ref{fig:ISMCTSUCT} reports the winning rate
	as a function of the ISUCT constant when ISMCTS plays as (a) the deck team and as (b) the hand team
	using 4000 iterations against the Greedy strategy.\footnote{Note that, ISMCTS needs four times more iterations before its performance stabilizes.}
The trends are similar to the ones reported for MCTS (Figure~\ref{fig:MCTSUCT}) 
    and the plots basically confirm the results for MCTS: the best ISUCT constant for PWL is 0.35 and it is 2
	for SD while RS still leads to the worst performance and thus
	it was dropped for the subsequent experiments. 
The performance for WL shows different behavior when playing as the deck team or the hand team for 
	the values 0.5, 0.7, and 1.0. However, when considering the average performance the three constants 
	lead to similar (not statistically different) values. Accordingly, for WL we selected the same
	constant used for MCTS, that is, 0.7.

\begin{figure*}[t]

\begin{tabular}{cc}
\parbox{\columnwidth}{
\centering

\includegraphics[width=\plotwidth]{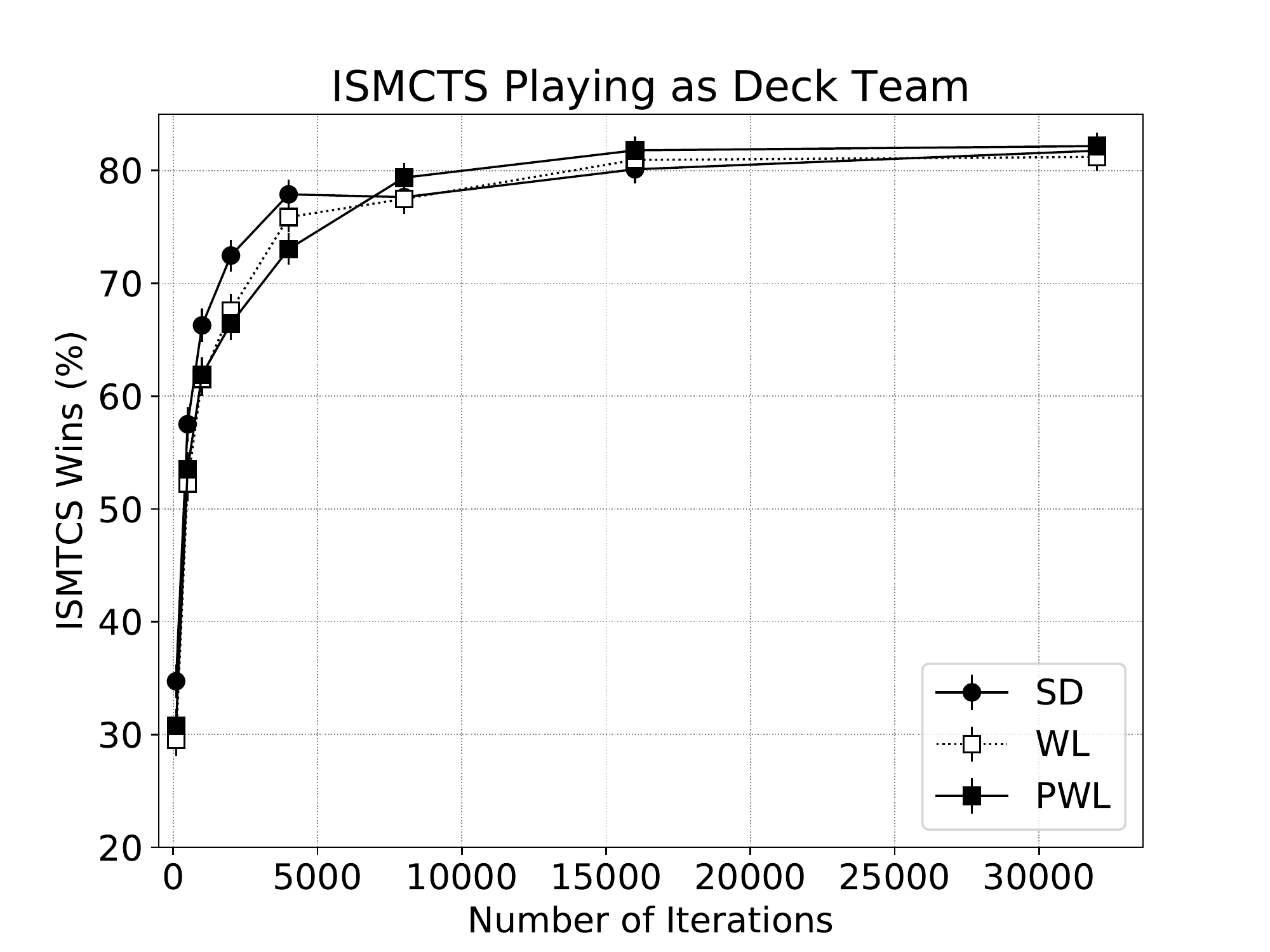}

\centerline{(a)}
} & 
\parbox{\columnwidth}{
\centering

\includegraphics[width=\plotwidth]{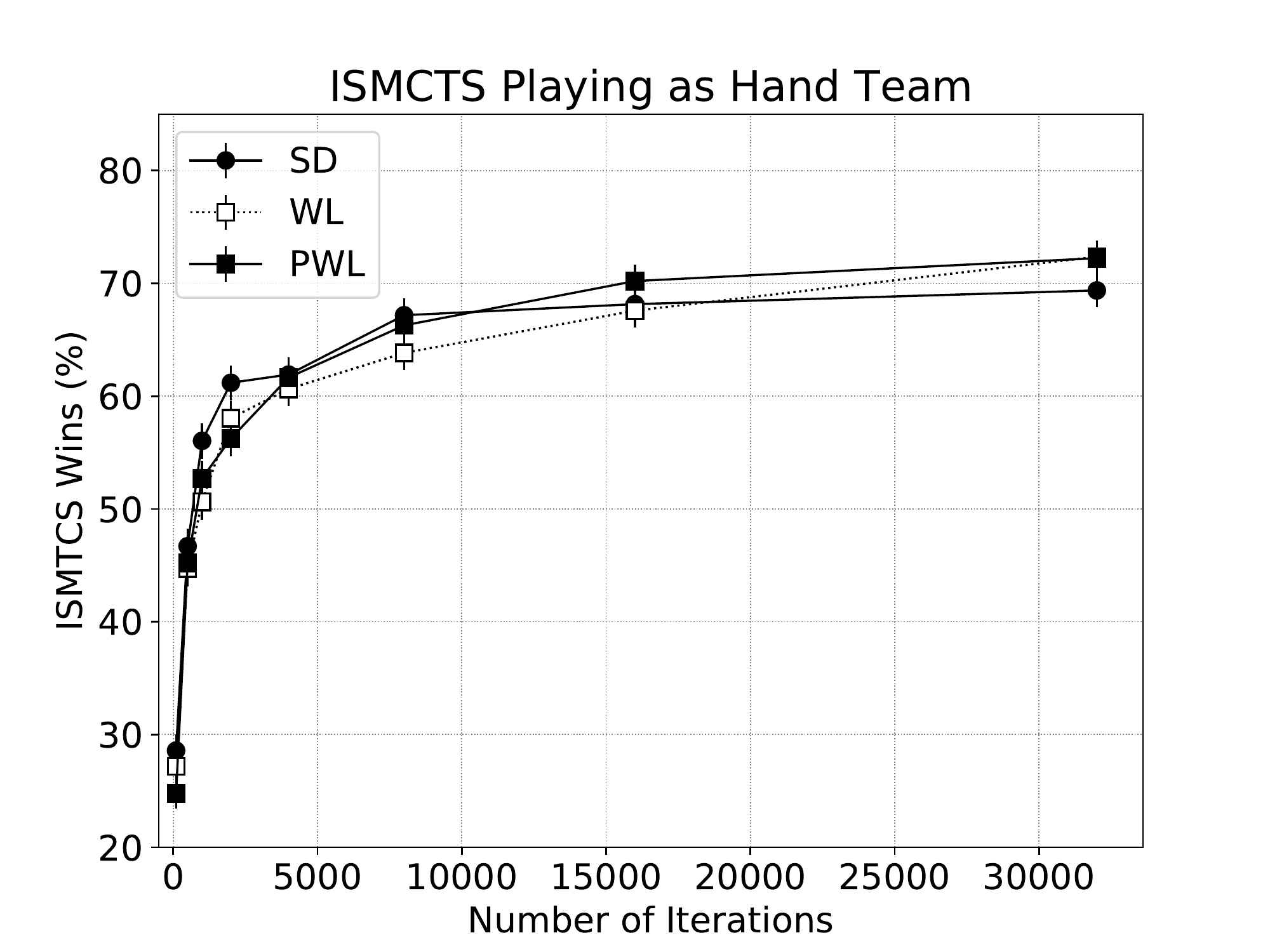}

\centerline{(b)}
}\\
\end{tabular}

\caption{Comparison of reward functions for ISMCTS. Winning rate as a function of the number of iterations when ISMCTS plays against the Greedy strategy as (a) the deck team and as (b) the hand team; bars report the standard error.}

\label{fig:ISMCTSrewards}

\end{figure*}

\begin{figure}
\centering
\includegraphics[width=\plotwidth]{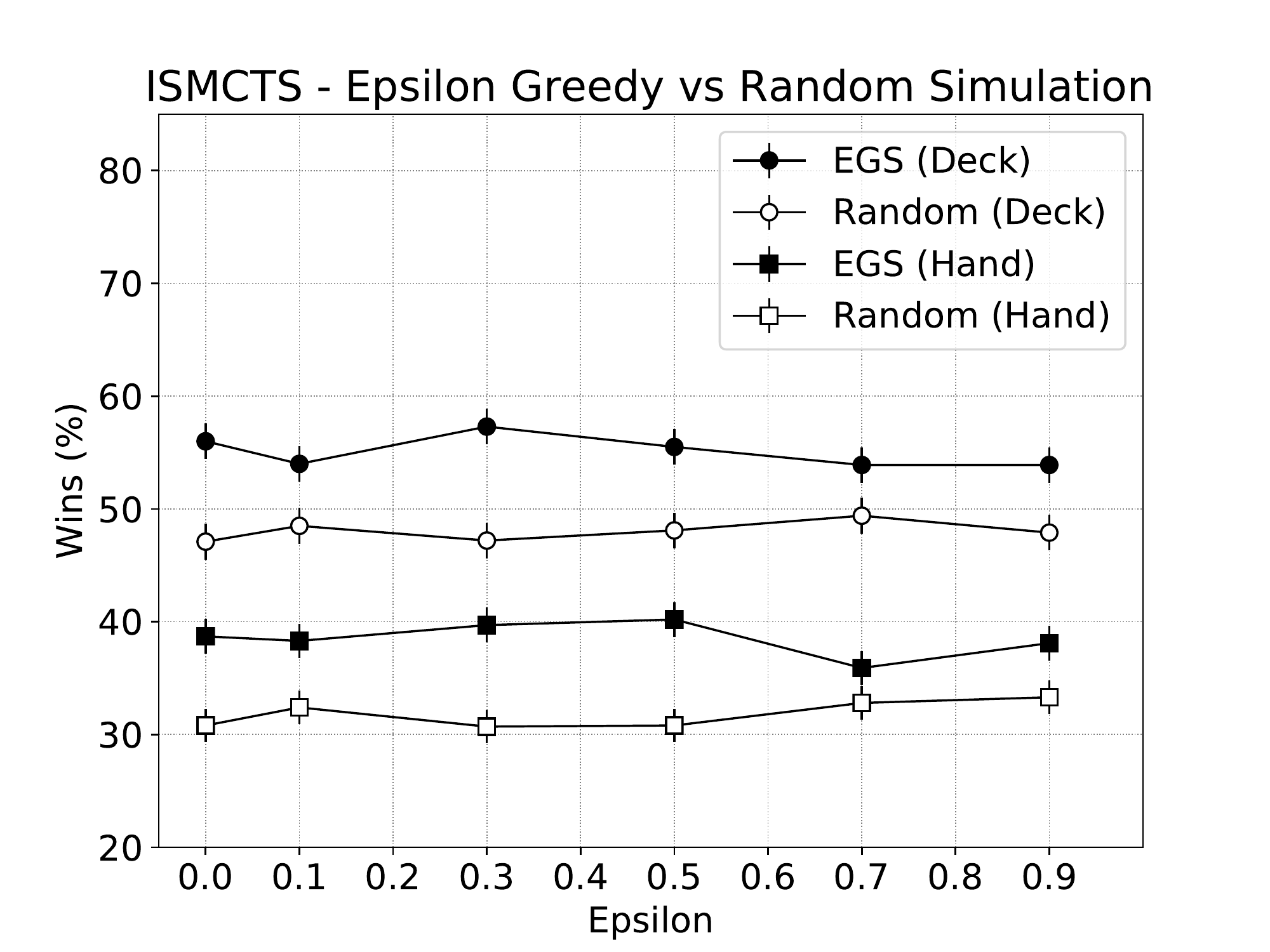}

%
%
%
%
%
%
%
\caption{Winning rate of ISMCTS using SD with a ISUCT constant of 2 as a function of the $\epsilon$ value used by the Epsilon-Greedy Simulation strategy when it plays against the same ISMCTS using random simulation. 
Bars report the standard error.
\label{fig:ISMCTSEGS}}
%
\centering

\includegraphics[width=\plotwidth]{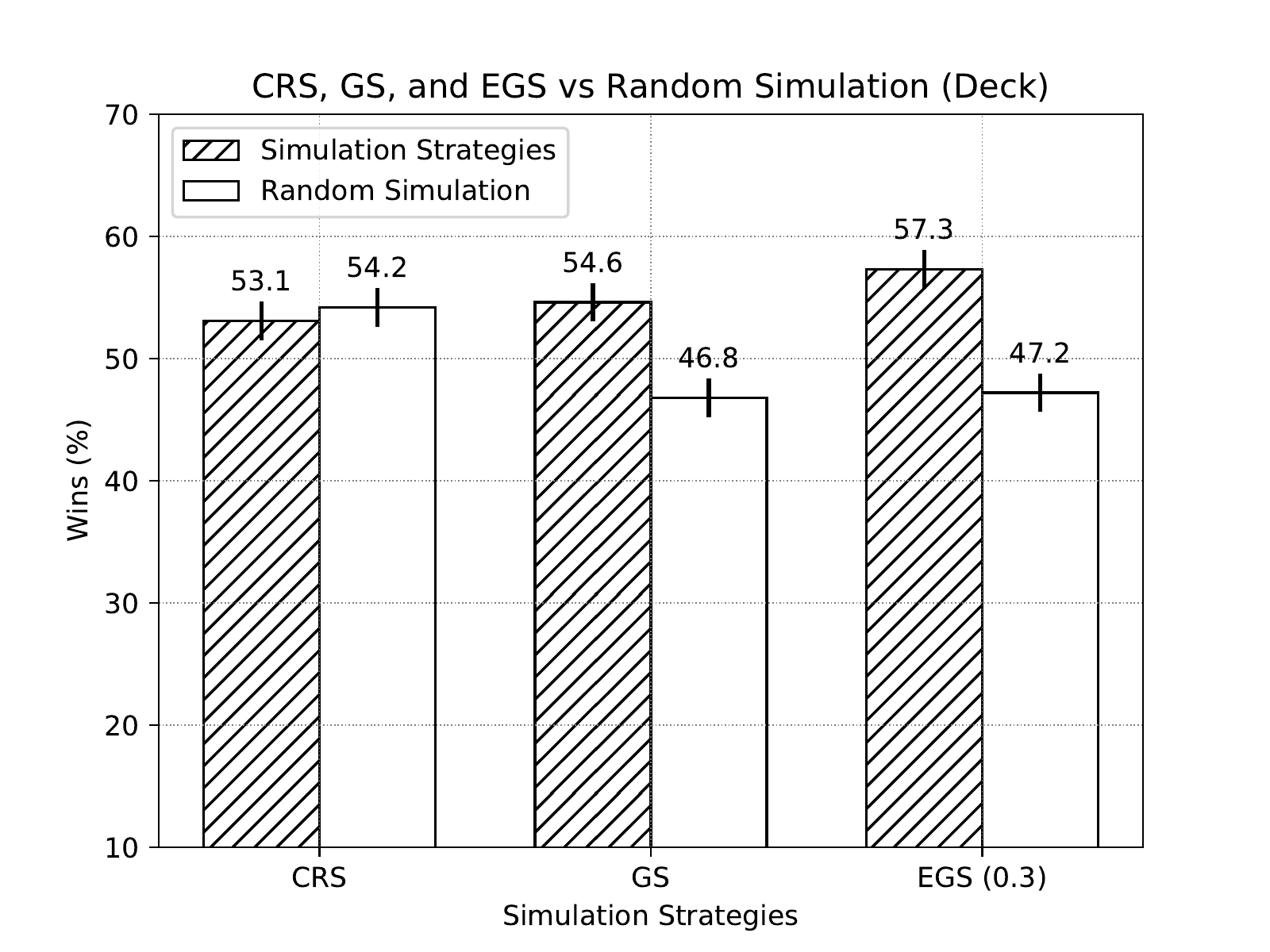}

\centerline{(a)}

\includegraphics[width=\plotwidth]{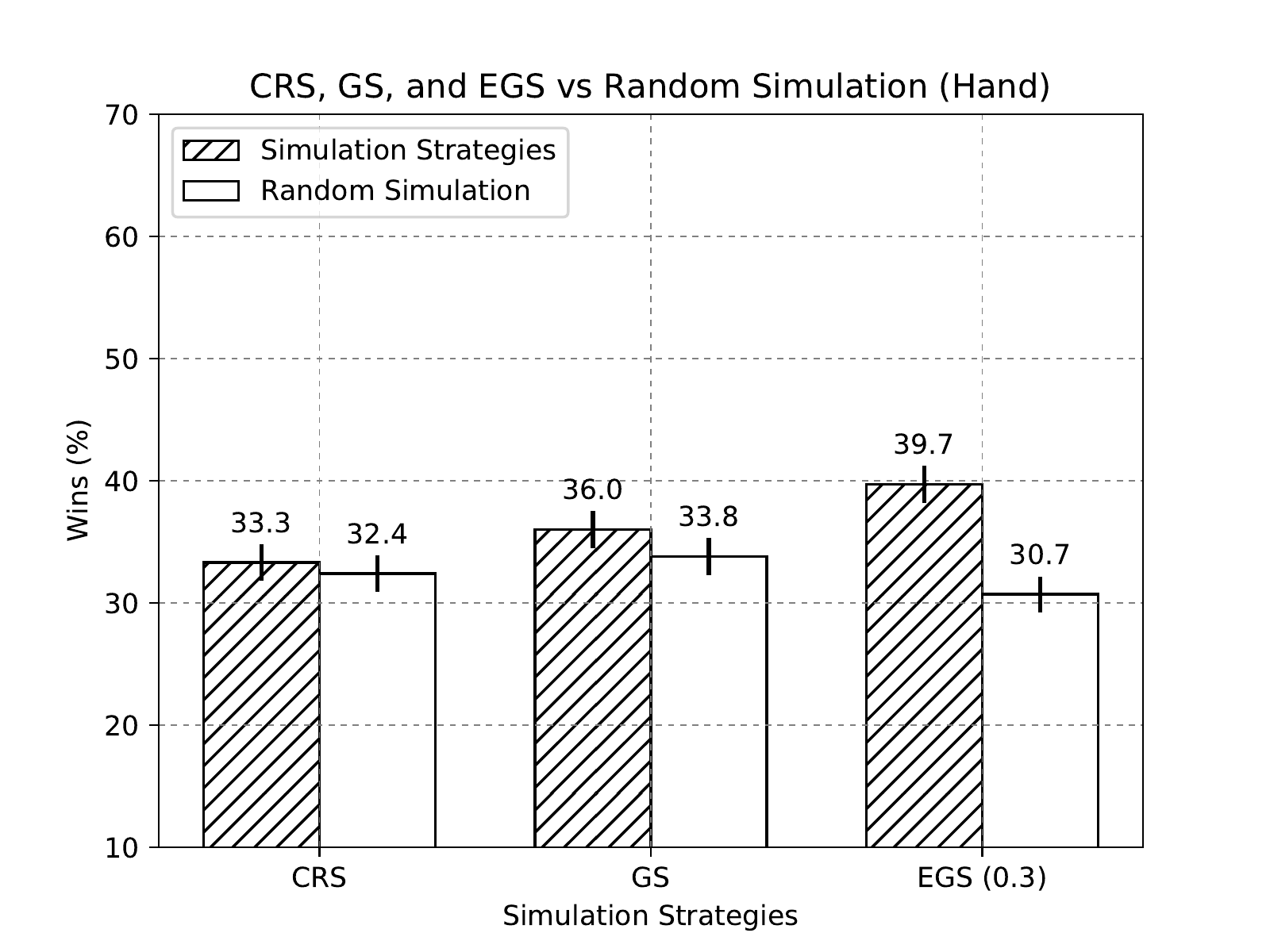}

\centerline{(b)}

\caption{Winning rate of ISMCTS using CRS, GS and EGS with an epsilon of 0.3 compared to ISMCTS using random simulation when playing as (a) the deck and (b) the hand team. Bars report the standard error.
\label{fig:ISMCTSSimulationStrategy}}
\end{figure}

\medskip\noindent\textbf{Reward Functions.}
Figure~\ref{fig:ISMCTSrewards} reports the winning rate as a function of the number of iterations when playing against the Greedy strategy.
The plots are similar to what reported for MCTS: as the number of iterations increases, the three reward
functions perform similarly and there is no statistically significant difference beyond 4000 iterations.
With 1000 iterations, SD performs significantly better than WL both as the deck team and as the hand team (p-value is 0.035 and 0.031 respectively); with 2000 iterations SD performs significantly better than WL when playing for the deck team; all the other differences in Figure~\ref{fig:ISMCTSrewards} are not statistically significant.
Furthermore, for all the reward functions, 
	the increase of performance beyond 4000 iterations becomes statistically significant when the number of iterations are at least quadrupled, e.g., from 4000 to 16000 or
	from 8000 to 32000.
Finally, the results show that ISMCTS with 4000 iterations outperforms the Greedy strategy by winning $66.6\% [63.68-69.52]$ of the matches as deck team (when using SD) and the $53.0\% [49.91-56.09]$ as hand team against the Greedy strategy.
Moreover, they suggest that ISMCTS can better exploits the advantages of playing for the deck team.

\medskip\noindent\textbf{Simulation Strategies.}
As we did with MCTS, we compared four different simulation strategies for ISMCTS:
	\emph{Random Simulation} (RS), \emph{Card Random Simulation} (CRS), 
	\emph{Greedy Simulation} (GS), and \emph{Epsilon-Greedy Simulation} (EGS).
As before, we used the standard random simulation strategy (RS) as the baseline to compare the improvement provided by the other three heuristics.
Figure~\ref{fig:ISMCTSEGS} compares the performance of ISMCTS using 4000 iterations, SD and EGS (solid dots) 
	with different value of $\epsilon$ (0.0, 0.1, 0.3, 0.5, 0.7, and 0.9) 
	against the same ISMCTS using RS (white dots).
EGS outperforms random simulation both when playing as the deck (circle dots) and as the hand team (square dots).
Compared to the results for MCTS (Figure~\ref{fig:MCTSEGS}), 
	the plots for ISMCTS are flattened and EGS performs only slightly better for 0.3 when playing as the deck team
	although the difference is not statistically significant when compared to the other values of $\epsilon$. 
Note that, plain greedy simulation ($\epsilon=0$) does not harm performance as it did for MCTS.
%
Figure~\ref{fig:ISMCTSSimulationStrategy} compares the performance of ISMCTS using plain random simulation against 
	\emph{Card Random Simulation} (CRS),
	\emph{Greedy Simulation} (GS), and \emph{Epsilon-Greedy Simulation} (EGS) with $\epsilon=0.3$ when playing as (a) the deck and (b) the hand team.
Greedy strategy performs surprisingly well when considering the results for MCTS (Figure~\ref{fig:MCTSSimulationStrategy}) while CRS still performs similarly to random simulation.
EGS outperforms plain random strategy both when playing as the hand and as the deck team; 
	it performs slightly better than GS when playing as the deck team and the hand team but the difference is not statistically significant
	(p-values are 0.224 and 0.088 respectively).
Overall, EGS results in the highest improvement over default random strategy, accordingly,
	we selected it with $\epsilon=0.3$ for the final tournament.

\begin{figure}[t]
\centering
\includegraphics[width=\plotwidth]{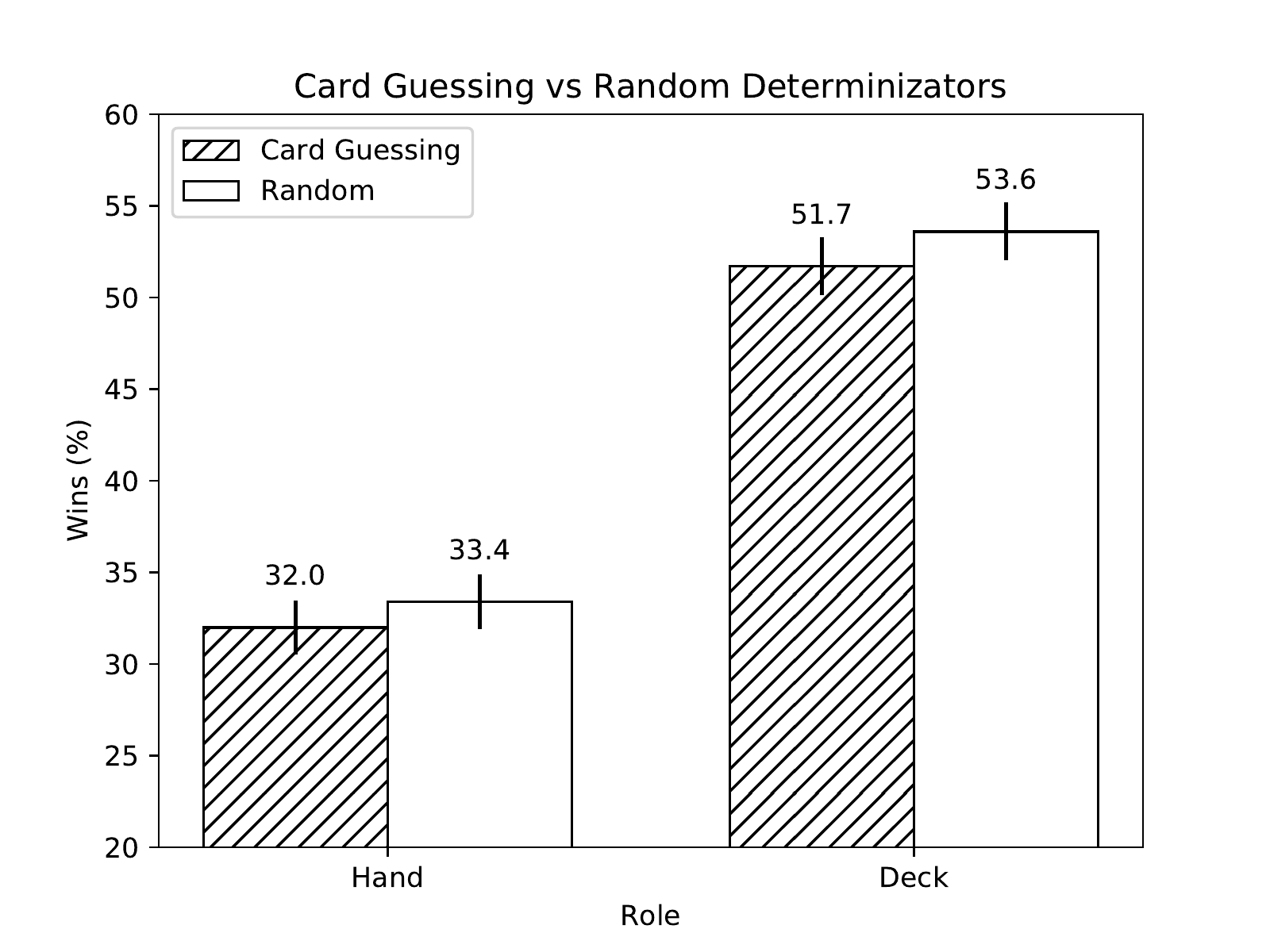}

\caption{Winning rates of the Card Guessing System determinizator (stripped bars) and the Random determinizator (white bars) 
		when they play against each other as the hand and the deck team. Bars report the standard error.
\label{fig:ISMCTSDeterminizatorsVsCS}}
\end{figure}

\medskip\noindent\textbf{Determinizators.}
We compared two methods to determinize the root information set at each iteration using the parameters selected in the previous experiments,
	that is, an ISUCT constant of 2.0, Score Difference reward function, and epsilon-Greedy Strategy with $\epsilon=0.3$.
The \emph{Random} determinizator \cite{DBLP:journals/tciaig/CowlingPW12} samples a state within the root information set, while the \emph{Cards Guessing System} (CGS) determinizator restricts the sample to the states in which each player holds the cards guessed by the cards guessing system (Section~\ref{sec:rulebasedai}).
Figure~\ref{fig:ISMCTSDeterminizatorsVsCS} compares Random and CGS determinizators, with ISMCTS playing both as hand team and deck team. Random determinizator performs slightly better than CGS but the difference is not 
statistically significant (p-value is 0.5046 and 0.3948 respectively). 
The analysis of specific matches showed that CGS allows ISMCTS to avoid moves that bring the opponents to an easy scopa, since with CGS can guess the cards that the opponents might hold. For this reason, we selected CGS as the determinizator for the final tournament.

\subsection{Final Tournament}
\label{sec:finalTournament}
At the end, we performed a tournament among 
	the random strategy and the best players selected with the previous experiments, namely:
(1) Chitarella-Saracino (CS), that resulted as the best rule-based player;
(2) Monte Carlo Tree Search (MCTS) using Scores Difference (SD) rewards, 
	an UCT constant equals to 2, the $\epsilon$-Greedy Simulation strategy with $\epsilon=0.3$, 
	and 1000 iterations;
(3) Information Set Monte Carlo Tree Search (ISMCTS) using Scores Difference (SD) rewards, 
	an ISUCT constant equals to 2, the $\epsilon$-Greedy Simulation strategy with $\epsilon=0.3$, 
	the \emph{Cards Guessing System} determinizator, and 4000 iterations. 
Note that we set the number of iterations to 1000 for MCTS and 4000 for ISMCTS
	since, as we noted before, a series of preliminary experiments showed 
	that is when MCTS and ISMCTS reach a plateau.
Table~\ref{tab:finalTournament} reports the results of the tournament, 
	and 	Table~\ref{tab:finalScoreboard} shows the final scoreboard.
Unsurprisingly, playing at random is the worst strategy:
	it loses the $84.8\%$ of times and wins only the $10.3\%$ of games, the majority of them against itself. MCTS confirms to be the best strategy winning $79.0\%$ of the games
	and loosing $12.6\%$ of games (mostly when playing against itself).
This result was also expected, 
	since MCTS has complete knowledge of the cards of all the other players.
The comparison between CS and ISMCTS is more interesting,
	since they play fair and have a partial knowledge of the game state.
ISMCTS outperforms CS by winning $55.8\%$ of the games and losing $34.1\%$ of games,
	whereas CS wins $41.7\%$ of the games and loses $47.9\%$ of the games.
Thus, ISMCTS proved to be stronger than an artificial intelligence 
	implementing the most advanced strategy for Scopone 
	that was firstly designed by expert players centuries ago and updated ever since. 
Finally, the results confirm that the deck team has an advantage over the hand team
	and that such advantage increases with the ability of the player.
In fact, if we consider the matches between the same artificial intelligence 
	(the diagonal values of Table~\ref{tab:finalTournament}), 
	we note that the winning rate of the hand team decreases as the player's strength increases: 
	$38.0\%$ of the random strategy, $38.1\%$ of CS, $34.7\%$ of ISMCTS, and $29.5\%$ of MCTS.

\begin{table*}[t]
\centering
\caption{Result of the final tournament. The artificial intelligence used by the hand team is listed on the left, the one used by the deck team is listed at the top.
(a) percentage of wins by the hand team;
(b) percentage of losses by the hand team; 
(c) percentage of ties.}
\label{tab:finalTournament}

\scriptsize

\begin{tabular}{|c||c|c|c|c|}
\hline
\multicolumn{5}{|c|}{\textbf{Winning rates of the hand team}} \\
\hline
$\nicefrac{\textbf{Hand Team}}{\textbf{Deck Team}}$& \textbf{Random} & \textbf{CS} & \textbf{MCTS} & \textbf{ISMCTS}\\\hline
\textbf{Random}	& 
$ 38.0 \% [35.9 - 41.0] $	& 
$ 3.5 \% [2.4 - 4.6] $		& 
$ 0.0 \% [0.0 - 0.0] $		&	
$ 0.3 \% [0.0 - 0.6] $	\\ \hline 
\textbf{CS}	& 
$ 91.1 \% [89.3 - 92.9] $			&	
$ 38.1 \% [35.1 - 41.1] $			&	
$ 2.4 \% [1.5 - 3.3] $		&	
$ 20.9 \% [18.4 - 23.4] $	\\\hline	
\textbf{MCTS} &
$ 99.0 \% [98.4 - 99.6] $			&	
$ 85.4 \% [83.2 - 87.6] $			&	
$ 29.5 \% [26.7 - 32.3] $			&	
$ 70.0 \% [67.2 - 72.8] $			\\\hline	
\textbf{ISMCTS} & 
$ 96.2 \% [95.0 - 97.4] $			&	
$ 56.3 \% [53.2 - 59.4] $			&	
$ 5.5 \% [4.1 - 6.9] $					&	
$ 34.7 \% [31.7 - 37.7] $			\\\hline	
\end{tabular}

\vskip.2cm
\centerline{(a)}
\vskip.2cm

\begin{tabular}{|c|c|c|c|c|}
\hline
\multicolumn{5}{|c|}{\textbf{Losing rates of the hand team}} \\
\hline
$\nicefrac{\textbf{Hand Team}}{\textbf{Deck Team}}$& \textbf{Random} & \textbf{CS} & \textbf{MCTS} & \textbf{ISMCTS}\\\hline
\textbf{Random}	& 
$ 49.5 \% [46.4 - 52.6] $	& 
$ 93.4 \% [91.9 - 94.9] $	& 
$ 99.8 \% [99.5 - 100.0]$	&	
$ 97.1 \% [96.1 - 98.1]$		\\ \hline 

\textbf{CS}	& 
$ 4.1 \% [2.9 - 5.3] $	&	
$ 46.8 \% [43.7 - 49.9] $	&	
$ 92.5 \% [90.9 - 94.1] $	&	
$ 68.0 \% [65.1 - 70.9] $	\\\hline	

\textbf{MCTS} &
$ 0.1 \% [0.0 - 0.3] $	&	
$ 7.7 \% [6.0 - 9.4] $	&	
$ 52.5 \% [49.4 - 55.6] $	&	
$ 18.7 \% [16.3 - 21.1] $	\\\hline	
\textbf{ISMCTS} & 
$ 1.1 \% [0.5 - 1.7] $	&	
$ 31.0 \% [28.1 - 33.9] $	&	
$ 84.6 \% [82.4 - 86.8] $	&	
$ 50.8 \% [47.7 - 53.9] $	\\\hline	
\end{tabular}

\vskip.2cm
\centerline{(b)}
\vskip.2cm

\begin{tabular}{|c|c|c|c|c|}
\hline
\multicolumn{5}{|c|}{\textbf{Tying rates}} \\
\hline
$\nicefrac{\textbf{Hand Team}}{\textbf{Deck Team}}$& \textbf{Random} & \textbf{CS} & \textbf{MCTS} & \textbf{ISMCTS}\\\hline
\textbf{Random}	& 
$ 12.5 \% [10.5 - 14.5] $	& 
$ 3.1 \% [2.0 - 4.2] $	& 
$ 0.2 \% [0.0 - 0.5] $	&	
$ 2.6 \% [1.6 - 3.6] $	\\ \hline 

\textbf{CS}	& 
$ 4.8 \% [3.5 - 6.1] $	&	
$ 15.1 \% [12.9 - 17.3] $	&	
$ 5.1 \% [3.7 - 6.5] $	&	
$ 11.1 \% [9.2 - 13.0] $	\\\hline	

\textbf{MCTS} &
$ 0.9 \% [0.3 - 1.5] $	&	
$ 6.9 \% [5.3 - 8.5] $	&	
$ 18.0 \% [15.6 - 20.4] $	&	
$ 11.3 \% [9.3 - 13.3] $		\\\hline	

\textbf{ISMCTS} & 
$ 2.7 \% [1.7 - 3.7] $	&	
$ 12.7 \% [10.6 - 14.8] $	&	
$ 9.9 \% [8.0 - 11.8] $	&	
$ 14.5 \% [12.3 - 16.7] $	\\\hline	
\end{tabular}

\vskip.2cm
\centerline{(c)}
\vskip.2cm
\end{table*}


%
%
%
\begin{table}
\centering
\caption{Percentage of wins, losses, and ties for each artificial intelligence during 
the  tournament.}
\label{tab:finalScoreboard}

\begin{center}
\begin{scriptsize}

\begin{tabular}{|c||c|c|c|}
\hline
\textbf{AI} & \textbf{Wins} & \textbf{Losses} & \textbf{Ties} \\
\hline
\hline
\textbf{Random} & $10.3\%$  			& $84.8\%$ 			& $4.9\%$    \\
							& $[9.0 - 11.7]$  		& $[83.2 - 86.4]$	& $[3.9 - 5.8]$    \\
\hline
\textbf{CS}			& $41.7\%$ 				& $47.9\%$ 			& $10.4\%$ \\
							& $[39.6 - 43.9]$ 	& $[45.7 - 50.1]$ 	& $[9.0 - 11.7]$ \\
\hline
\textbf{MCTS}		& $79.0\%$ 			& $12.6\%$ 			& $8.3\%$\\
							& $[77.3 - 80.8]$ 	& $[11.2 - 14.1]$	& $[7.1 - 9.5]$\\
\hline
\textbf{ISMCTS}	& $55.8\%$ 			& $34.1\%$ 				& $10.1\%$\\
							& $[53.6 - 57.9]$ 	& $[32.1 - 36.2]$ 	& $[8.8 - 11.4]$\\
\hline
\end{tabular}

\end{scriptsize}
\end{center}
\end{table}

\begin{figure}[t]
\centering
\includegraphics[width=\plotwidth]{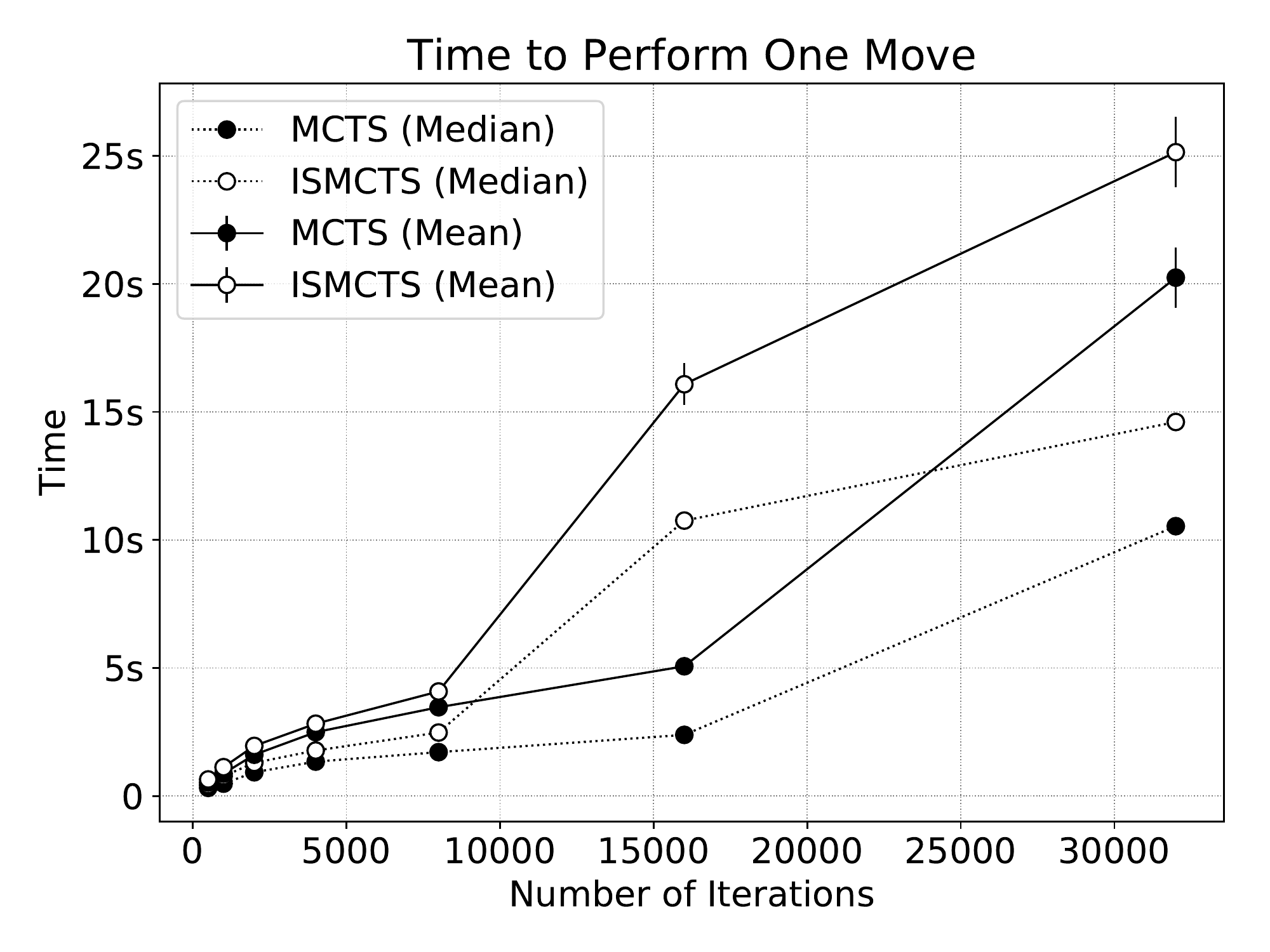}
\caption{Time to perform one move using MCTS (solid marker) and ISMCTS (white marker)
	as a function of the number of iterations: 
	average (solid lines); median (dotted lines). Bars report the standard error.}
\label{fig:timing}
\end{figure}

\medskip\noindent\textbf{Increasing the Number of Iterations.}
ISMCTS generally needs more iterations than MCTS \cite{DBLP:journals/tciaig/CowlingPW12}, 
  accordingly, we run an experiment to compare the performance of ISMCTS using 32000 iterations
  against the players used in the previous experiment.
Table~\ref{tab:final_ismcts32} reports the percentage of matches that the new version of ISMCTS  won, lost, and tied.
As expected, the higher number of iterations improves ISMCTS performance.
The percentage of wins against Random, CS, and MCTS increased  
	(from 96.2\% to 96.8\% against Random, from 56.3\% to 64.8\% against CS, and from 5.5\% to 13.2\% against MCTS) and the result is statistically significant for CS and MCTS.
Similarly, the percentage of lost matches decreased 
	(from 1.1\% to 0.8\% against Random, from 31.0\% to 23.3\% against CS, and from 84.6\% to 77.7\% against MCTS) and the results are statistically significant for CS and MCTS.
When comparing the performance of ISMCTS using 32000 with that of ISMCTS using 4000 iterations, 
	the higher number of iterations results in more wins (46.1\% against 34.7\%) and
	fewer losses (40.2\% instead of 50.8\%).
However, this increase in performance has a cost. Figure~\ref{fig:timing}
	plots the average time and the median time needed to select an action using MCTS and ISMCTS as a function of the number of iterations; 
	the averages and medians are computed over 1000 measurements that were performed with a Intel Core i5 3.2Ghz with 16 GB.
As the number of iterations increases,
	the average time needed to perform a move dramatically increases 
	and rapidly becomes infeasible for interacting with players. 
In fact, our implementation of MCTS and ISMCTS can take an average of 20 seconds to 
	select a move using 32000 iterations and 
	as the number of iterations increases, the variance increases
	(see the larger bars for higher numbers of iterations). 	
We profiled our implementation (based on the Mono\footnote{\url{http://www.mono-project.com/}} C\#)
	and noted that most of the time is spent to compute the available
	admissible moves given a player's cards and the cards on the table.
Our analysis suggests
	that the variance increase is mainly due to memory management and optimization overhead introduced by the C\# runtime environment.
	
%
%
%
%

\begin{table}
\caption{Performance of ISMCTS using 32000 iterations playing against
the four strategies used in the previous experiment (Random, CS, MCTS, and ISMCTS using 4000 iterations) reported as the percentage of wins, losses, and ties.}

\begin{center}
\begin{scriptsize}

\begin{tabular}{|c||c|c|c|}
\hline
 & \multicolumn{3}{c|}{\textbf{ISMCTS (32000 Iterations)}} \\ \cline{2-4}
\textbf{AI}				& \textbf{Wins} 				& \textbf{Losses} 			& \textbf{Ties} \\ \hline\hline
\textbf{Random}					& $ 96.8\%$						& $0.8\%$ 						& $2.4\%$ \\
											& $ [96.1 - 97.6] $				& $[0.4 - 1.2] $ 				& $[1.7 - 3.0] $ \\ \hline
\textbf{CS}							& $ 64.8\%$						& $ 23.3\%$					& $11.8 \%$ \\ 
											& $ [62.7 - 66.9] $				& $ [21.5 - 25.2] $			& $[10.4 - 13.3] $ \\ \hline
\textbf{MCTS}						& $ 13.2\%$							& $ 77.7 \%$					& $ 9.2  \%$ \\ 
											& $ [11.7 - 14.6] $				& $ [75.9 - 79.5] $			& $ [7.9 - 10.4] $ \\ \hline
\textbf{ISMCTS}					& $ 46.1\%$							& $ 40.2\%$					& $ 13.7\%$ \\ 
											& $ [43.9 - 48.2] $				& $ [38.1 - 42.4] $			& $ [12.2 - 15.2] $ \\ \hline
\end{tabular}

\end{scriptsize}
\end{center}
\label{tab:final_ismcts32}
\end{table}

\medskip\noindent\textbf{Human Players.} We performed a final experiment with human players using a simple prototype we developed with 
Unity.\footnote{\url{http://www.unity3d.com}}
Our goal was to get a preliminary evaluation of the performance of the strategies we developed when they faced human players. 
We asked amateur Scopone players to play against an artificial player randomly selected from (i) Greedy strategy, (ii) Chitarella-Saracino (CS), 
	(iii) the cheating MCTS using 1000 iterations; (iv) ISMCTS using 1000 iterations; and (v) ISMCTS using 4000 iterations.
At the start of a match, 
	one of the five strategies was randomly selected and assigned both to the human player's teammate and to the opponent players;
	then, the starting player would be randomly selected.
Note that, human players did not know how the artificial players were implemented nor the strength of the players they were facing and could not guess it
	from the time taken for selecting a move since we added a random delay to the move selection procedure.
	
We collected data of 218 matches involving 32 people but after cleaning some incomplete data 
	we ended up with 105 matches recorded (21 for each one of the five artificial players 
	included in the prototype). 
The data show that human players won 30.5\% of the matches, 
	tied 12.4\% of the matches, and lost the remaining 57.1\%.
Table~\ref{tab:humans} reports the percentage of matches that human players won, 
	lost and tied for each strategy considered. 
Unsurprisingly, the cheating MCTS player won most of the matches (90.5\%).
Human players found it easier to play against the basic Greedy strategy
	than the slightly more advanced Chitarella-Saracino (CS), 
	winning 47.6\% of the matches against Greedy but only 42.9\% against CS;
	human players also lost more matches against CS than Greedy.
ISMCTS is a more challenging opponent for human players
	that lost only 33.3\% of the matches with 1000 iterations and 
	23.8\% with 4000 iterations. 
Accordingly, ISMCTS outperforms all the fair players (Greedy and CS) 
	the difference is statistically significant at a 95\% confidence level 
	for Greedy (p-value is 0.0212) and it is borderline for CS (p-value is 0.0623).\footnote{Note that given the few data points the confidence interval are rather broad and these are the only statistically significant differences together to the ones involving MCTS.}
We allowed players to submit comments about their experience. Some reported that they were puzzled by their teammate's behavior;
	when we checked the game logs we noted that these comments were mainly related to advanced strategies (MCTS and ISMCTS).
In our opinion, this suggests that amateur player can connect to more intuitive strategies (like Greedy and CS) but might be unable to grasp 
	an advanced strategy that does not follow what is perceived as the traditional way to play but simply focuses on optimizing the final outcome.

%
%
%

\begin{table}[t]
\caption{Percentage of matches that human players won, lost, or tied when playing with a teammate and two opponents based on the same strategy.}

\begin{center}
\begin{scriptsize}

\begin{tabular}{|c||c|c|c|}\hline
    & \multicolumn{3}{c|}{\textbf{Human Players}} \\\cline{2-4}
\textbf{AI} 			& \textbf{Wins} 	& \textbf{Losses} & \textbf{Ties}			\\ \hline\hline
\textbf{Greedy}					& $47.6\%$							& $38.1\%$								& $14.3\%$							\\		
											& $[26.3-69.0]$					& $[17.3 - 58.9]$					& $[0.0 - 29.3]$					\\\hline
\textbf{CS}		 					& $42.9\%$							& $47.6\%$								& $9.5\%$							\\
						 					& $[21.7 - 64.0]$				& $[26.3 - 69.0]$					& $[0.0 - 22.1]$					\\\hline
\textbf{MCTS}						& $4.8\%$							& $90.5\%$								& $4.8\%$							\\
											& $[0.0 - 13.9]$					& $[77.9 - 100.0]$					& $[0.0 - 13.9]$					\\\hline
\textbf{ISMCTS(1000)}		& $33.3\%$							& $52.4\%$								& $14.3\%$							\\
											& $[13.2 - 53.5]$				& $[31.0 - 73.7]$						& $[0.0 - 29.3]$					\\\hline
\textbf{ISMCTS(4000)}		& $23.8\%$							& $57.1\%$								& $19.0\%$							\\
											& $[5.6 - 42.0]$					& $[36.0 - 78.3]$					& $[2.3 - 35.8]$					\\\hline
\end{tabular}
\end{scriptsize}
\end{center}
\label{tab:humans}
\end{table}

\section{Conclusions}
\label{sec:conclusions}
We investigated the design of a competitive artificial intelligence for \emph{Scopone}, a traditional and very popular Italian card game 
	whose origins date back to 1700s.
We developed three rule-based artificial players corresponding to three well-known strategies:
	the \emph{Greedy} player (GS), implementing the strategy taught to beginners;
	a player implementing the rules of \emph{Chitarrella-Saracino} (CS) described in the most important and historical Scopone  books
	\cite{chitarrella2002regole,saracino2011scopone};
	and a player implementing the rules of \emph{Cicuti-Guardamagna} (CG), the second most important strategy book \cite{cicuti1978segreti} 
	that extends \cite{chitarrella2002regole,saracino2011scopone}.
We also developed a players based on MCTS and one based on ISMCTS.
The former requires the full knowledge of the opponent cards and thus, 
	by implementing a cheating player, provided an upper bound of the performance achievable using MCTS approaches.
The latter  implements a fair player that only knows what has been played and tries to guess (via determinization)
	the opponents' cards.
We used a set of experiments to determine the best configuration for MCTS and ISMCTS and 
	then performed a tournament involving the best rule-based player (CS), MCTS, and ISMCTS.
Our results show that, as expected, the cheating MCTS player outperforms all the other strategies.
ISMCTS  outperforms CS both when playing as the hand and as the deck team.
Interestingly, the rules of \emph{Chitarrella-Saracino}, which are still considered the most important expert guide for the game, 
	provide a very good strategy leading the hand team to victory around the 20\% 	of the time against ISMCTS 
	and the advantaged deck team to victory around the 31\% of the time.
At the end, we performed an experiment with human players to get a preliminary evaluation of how well our artificial players perform when
	facing human opponents. Overall, human players won 30.5\% of the matches, tied 12.4\% of them and lost the remaining 57.1\%; 
	they won more matches against simpler strategies (47.6\% against Greedy and 42.9\% against CS), 
	lost or tied almost all the matches against the cheating MCTS players, 
	and won only the 23.8\% of the matches against the fair ISMCTS player. 
Thus, it confirms that ISMCTS outperforms the most studied and well-established strategies for Scopone.
	
There are several areas for potential future research. 
Following the footsteps of Whitehouse et al. \cite{spades}, we plan to develop a reliable implementation of Scopone for mobile platforms since most of the available apps focus on multiplayer features and provide rather basic single player experiences.
Our results show that ISMCTS is a challenging opponent to human players
	but our analysis also show that our current ISMCTS might be too inefficient, accordingly, 
	we plan to work on the optimization of our implementation and investigate well-known MCTS enhancements 
	(like AMAF and RAVE \cite{DBLP:journals/tciaig/BrownePWLCRTPSC12}) and the more recent one that constraints memory usage \cite{DBLP:conf/aiide/PowleyCW17}. 
Finally, since in Scopone players cannot exchange any sort of information, 
		player modeling techniques \cite{DBLP:journals/corr/Walton-RiversWB17,holmgard2015mcts,DBLP:conf/dagstuhl/YannakakisSLA13} might be introduced
		to improve the card guessing module that is used by rule-based AIs and for ISMCTS determinization.

\section*{Acknowledgement}
Stefano and Pier Luca wish to thank the three anonymous reviewers for their invaluable comments and the people who volunteered to play the game.


\end{document}